\documentclass{article}

\usepackage[preprint,nonatbib]{neurips_2021}

\PassOptionsToPackage{numbers,sort,comma,compress}{natbib}
\usepackage{natbib}

\usepackage{graphicx,xcolor}
\graphicspath{{images/}}
\usepackage[list=true]{subcaption}

\usepackage{mathtools}

\definecolor{light-blue}{rgb}{0.6,0.6,1}
\definecolor{orange}{rgb}{1,0.58,0	}
\definecolor{purplemountainmajesty}{rgb}{0.59, 0.47, 0.71}
\definecolor{applegreen}{rgb}{0.55, 0.71, 0.0}
\definecolor{hanpurple}{rgb}{0.32, 0.09, 0.98}
\definecolor{green(ryb)}{rgb}{0.4, 0.69, 0.2}
\definecolor{forestgreen(web)}{rgb}{0.13, 0.55, 0.13}
\definecolor{amethyst}{rgb}{0.6, 0.4, 0.8}
\definecolor{darkbyzantium}{rgb}{0.36, 0.22, 0.33}
\definecolor{darkslateblue}{rgb}{0.28, 0.24, 0.55}
\definecolor{darkblue}{rgb}{0.0, 0.0, 0.55}

\usepackage{multirow}
\usepackage[rightcaption]{sidecap}
\usepackage{stmaryrd}
\usepackage{array}
\usepackage{floatrow}
\usepackage{amsmath}
\usepackage{pifont}
\usepackage{graphicx} 

\usepackage{wrapfig}

\newcommand{\w}{\mathbf{\theta}}

\newcommand{\e}{\mathbf{e}}

\newcommand{\delw}{\delta \mathbf{\theta}}

\newcommand{\hess}{\mathbf{H}}
\newcommand{\fisher}{\mathbf{F}}

\newcommand{\hessinv}{\mathbf{\widehat{\fisher}}^{-1}}

\renewcommand{\vec}[1]{\mathbf{#1}}

\usepackage{amsmath,amsfonts,bm}
\usepackage{amssymb}

\def\1{\bm{1}}

\def\vb{{\mathbf{b}}}

\def\vg{{\mathbf{g}}}

\def\vq{{\mathbf{q}}}

\def\vv{{\mathbf{v}}}
\def\vw{{\mathbf{\theta}}}
\def\vx{{\mathbf{x}}}
\def\vy{{\mathbf{y}}}

\providecommand{\1}{\mathbf{1}}

\newcommand{\R}{\mathbb{R}}

\usepackage[utf8]{inputenc} 
\usepackage[T1]{fontenc}    
\usepackage{hyperref}       
\usepackage{url}            
\usepackage{booktabs}       
\usepackage{amsfonts}       
\usepackage{nicefrac}       
\usepackage{microtype}      
\usepackage{xcolor}         

\usepackage{algorithm}
\usepackage{algorithmic}
\usepackage{enumitem} 

\usepackage{amsmath,amssymb,amsthm}

\newtheorem{theorem}{Theorem}

\usepackage{hhline} 
\usepackage{float}
\usepackage{pgf}
\usepackage{printlen}
\usepackage{subcaption}
\usepackage{caption}
\newcommand{\parhead}[1]{ \noindent {\bfseries\boldmath\ignorespaces #1}\hskip 0.9em plus 0.3em minus 0.3em}

 \allowdisplaybreaks
 \AtBeginDocument{%
 \abovedisplayskip=4pt minus 1pt
 \abovedisplayshortskip=2pt plus 1pt
 \belowdisplayskip=4pt minus 1pt
 \belowdisplayshortskip=2pt plus 1pt
}
\setitemize{itemsep=0mm, leftmargin=5mm, topsep=1mm}
\setenumerate{itemsep=0mm, leftmargin=5mm, topsep=1mm}
\usepackage[noindentafter]{titlesec}
\titlespacing\section{0pt}{6pt plus 0pt minus 1pt}{2pt plus 0pt minus 1pt}
\titlespacing\subsection{0pt}{4pt plus 0pt minus 1pt}{1pt plus 1pt minus 1pt}
\titlespacing\subsubsection{0pt}{4pt plus 0pt minus 1pt}{1pt plus 1pt minus 1pt}

\newcommand{\idlow}[1]{\mathord{\mathcode`\-="702D\it #1\mathcode`\-="2200}}

\title{M-FAC: Efficient Matrix-Free Approximations of Second-Order Information}

\author{%
  Elias Frantar \\
  IST Austria \\
  \texttt{elias.frantar@ist.ac.at} \\
  \And
  Eldar Kurtic \\
  IST Austria \\
  \texttt{eldar.kurtic@ist.ac.at} \\
  \And
  Dan Alistarh \\
  { IST Austria \& Neural Magic}\\
  \texttt{dan.alistarh@ist.ac.at}
}

\begin{document}

\maketitle

\begin{abstract}
Efficiently approximating local curvature information of the loss function is a key tool for optimization and compression of deep neural networks. 
Yet, most existing methods to approximate second-order information have high computational or storage costs, which limits their practicality. 
In this work, we investigate matrix-free, linear-time approaches for estimating Inverse-Hessian Vector Products (IHVPs) for the case when the Hessian can be approximated as a sum of rank-one matrices, as in the classic approximation of the Hessian by the empirical Fisher matrix. 
We propose two new algorithms: the first is tailored towards network compression and can compute the IHVP for dimension $d$, if the Hessian is given as a sum of $m$ rank-one matrices, using $O(dm^2)$ precomputation, $O(dm)$ cost for computing the IHVP, and query cost $O(m)$ for any single element of the inverse Hessian. 
The second algorithm targets an optimization setting, where we wish to compute the product between the inverse Hessian, estimated over a sliding  window of optimization steps, and a given gradient direction, as required for preconditioned SGD. 
We give an algorithm with cost $O(dm + m^2)$ for computing the IHVP and $O(dm + m^3)$ for adding or removing any gradient from the sliding window.
These two algorithms yield state-of-the-art results for network pruning and optimization with lower computational overhead relative to existing second-order methods. 
Implementations are available at~\cite{Code} and~\cite{NMCode}.
\end{abstract}

\addtocontents{toc}{\protect\setcounter{tocdepth}{0}}

\section{Introduction}

Given the recent success and increasing impact of deep learning, there has been significant work on improving the fundamental technical tools underpinning its progress. 
One such tool is the ability to estimate the local geometry of the loss function for deep models, which often comes in the form of estimates for second-order (Hessian) information. 
Such information is critical in several settings, such as neural network optimization and pruning. 

Directly using Hessian information in the context of deep learning is infeasible: for example, just storing the Hessian matrix for the standard ResNet50 model~\cite{2016-he} would occupy 2.5 Petabytes.
These constraints have inspired significant work on efficient numerical approximations of the Hessian for deep neural networks, such as the line of work on the K-FAC approximation~\cite{2015-kingma, 2015-martens, 2016-ba, 2019-zheng}, or efficient block-wise approximations~\cite{yao2020adahessian, 2020-singh}. 
One of the classic approaches, which we focus on in this paper, is the \emph{empirical Fisher}~\citep{1992-hassibi, 1998-amari, amari2016information} approximation to the Hessian, written as:
\begin{equation}
\label{eq:empirical-fisher}
    \hess \simeq \widehat{\fisher} = \frac{1}{N} \sum_{i=1}^N \nabla \ell_i \cdot \nabla \ell_i^{\top},
\end{equation}
where $N$ is the number of samples, $\nabla \ell_i$ denotes the gradient w.r.t. the $i$th sample at the given point, and $\cdot$ is the outer product of individual gradients, which we view as column vectors. 

The empirical Fisher approximation is fairly standard, e.g.~\cite{1992-hassibi, 1998-amari, amari2016information}, and has been recognized to be useful in a variety of practical settings where exact estimation of the Hessian or of the true Fisher information matrix is not feasible.  
At the same time, there is still active research in the community on the conditions for its applicability~\cite{2014-martens, 2019-kunstner, 2020-singh, thomas2020interplay}. 
A useful property of this approximation is that it also allows to estimate the \emph{inverse} of the Hessian, which is essential in many applications.  

Specifically, the fact that the empirical Fisher can be written as a sum of rank-one matrices allows the use of the Woodbury-Sherman-Morrison inversion formula~\cite{woodbury1950inverting} to exactly compute its inverse, by recursively integrating terms corresponding to each gradient into the inverse. (Please see Equation~\ref{eq:woodfisher-inv} for an exact derivation.) 
This approach was independently proposed by~\cite{1992-hassibi, 1998-amari}, for pruning and optimization, respectively, where it was validated on small networks, with hundreds of weights. 

The idea was adapted to deep neural networks (DNNs) by~\cite{2020-singh}, through approximation of the inverse in \emph{small diagonal blocks}. 
The authors show improved approximation quality for the Hessian inverse relative to a simple diagonal approximation, and that this leads to state-of-the-art pruning results in terms of accuracy. 
Yet, this approach is limited by the block-wise approximation: for block size $B$ and dimension $d$, it requires $\Theta(Bdm)$ time to recursively build the block-wise Fisher approximation using $m$ gradients, and $\Theta(Bd)$ time and memory for computing the Inverse-Hessian-Vector-Products (IHVPs) necessary for estimating pruning statistics. 
Clearly, the ideal $B = d$ case is still intractable at scale, and generally it is still unknown whether efficient algorithms are possible for computing IHVPs in this context.

\parhead{Contribution.} We address this question by introducing two efficient algorithms for computing IHVPs under the empirical Fisher approximation, with computational and storage costs that are \emph{linear} in the dimension $d$ of the model, without the need for block-wise approximations, assuming that the number of samples $m$ in the approximation is constant. 
Concretely, we provide exact \emph{matrix-free} algorithms to compute products of the form $\widehat{\fisher}^{-1} \vv,$
where $\widehat{\fisher}^{-1}$ is the inverse empirical Fisher, and $\vv$ is an arbitrary vector.
We show that these algorithms can be implemented efficiently, and that they can match or improve the state-of-the-art results for both neural network pruning and optimization. 

\parhead{The Static Algorithm.} 
Our first algorithm assumes a \emph{static} scenario, which is standard in neural network pruning: 
we are given a fully-trained model $\theta^{\star}$, 
for which we wish to estimate IHVPs and diagonal elements of the inverse Hessian, in order to determine the ``optimal'' pruning update using e.g. the Optimal Brain Surgeon (OBS) framework~\citep{1990-lecun, 1992-hassibi}.  For this, we first compute $m$ gradients at $\theta^{\star}$, which we will use to estimate IHVPs via the empirical Fisher and compute pruning statistics.  

The main idea is that, since we only wish to compute products between the inverse and an arbitrary vector (IHVPs), 
we can rewrite the Woodbury recursion such that we work \emph{exclusively with individual vectors and scalars}, and never with full $d \times d$ or $d \times B$ matrices. 
Given model dimension $d$ and $m$ gradients, each defining a rank-one component of the empirical Fisher, the algorithm uses 
$O(dm^2)$ pre-computation time, and will have $O(dm)$ cost for \emph{exactly} computing the IHVP. 
Further, we can specialize the algorithm to directly query elements of the Hessian inverse, at a cost of $O(m)$ time per element. 
This provides efficient, linear-in-$d$ implementations for all operations required by the OBS pruning framework. Finally, we note that the static algorithm can also be applied in a block-wise manner without any change in the total compute and memory costs.

\parhead{The Dynamic Algorithm.} 
Our main contribution is in extending this idea to preconditioned SGD optimization, i.e. to precondition stochastic gradients by our estimate of the inverse Hessian. 
We start from the classic idea of bootstrapping the approximation by leveraging previous gradients: the preconditioner at time $t$ is built from gradients obtained during a ``sliding window''  over the last $\Delta \geq 1$ optimization steps.  
This requires a \emph{dynamic} representation, allowing addition and removal of gradients \emph{without full recomputation} of second-order statistics.

We show that this can be achieved, with approximately $O(dm)$ time and space complexity. 
The key idea is that, for any ordered set $(\nabla \ell_{j})_{j = 1}^m$ of $m$ gradients, and any vector $\vv$, it is possible to represent the corresponding IHVP estimate $\widehat{\fisher}^{-1}_m \vv$ as a linear combination of terms corresponding to individual gradients $\nabla \ell_{j}$ and $\vv$, of the form 
$\widehat{\fisher}^{-1}_m \vv = \lambda^{-1} \vv - \sum^m_{j=1} c^m_j \nabla \ell_{j}, \textnormal{ where $\lambda > 0$ is a dampening constant.}$
    
Crucially, we ensure that the coefficients $c^m_j$ can be computed just via dot products $\nabla \ell_{i}^{\top} \nabla \ell_{j}$, where $i \leq j$ in the ordering, and $\nabla \ell_{j}^{\top} \vv$. Then, to replace a given gradient $\nabla \ell_{i}$ from this representation, we just have to compute $m$ scalar products with the new gradient vector (as well as update some intermediate information). Hence, the entire update operation has computational cost $O(dm + m^3)$ for replacing any gradient in the sliding window, and $O(dm + m^2)$ for computing the IHVP.

\parhead{Implementation and Experiments.}
We provide efficient vectorized implementations for the above algorithms, called \emph{M-FAC}, for \emph{Matrix-Free Approximate Curvature}. Specifically, M-FAC consists of Pytorch~\cite{paszke2019pytorch} implementations of a pruning and optimization library.
Our implementation introduces several additional optimizations, in particular GPU acceleration via custom CUDA kernels, and minimizes the cost of memory transfer between the GPU and main memory via memory paging. 

For pruning, our implementation provides order-of-magnitude improvements over the block-wise approximation of~\cite{2020-singh} for classic benchmarks such as pruning ResNet50 and MobileNet on the ImageNet dataset. This allows us to obtain more accurate sparse models by exploring higher parameter settings and increasing the total number of pruning steps, while remaining practical in terms of memory and compute even for larger models. What is more, our preconditioned SGD (even without momentum) can be competitive in terms of validation accuracy with state-of-the-art optimizers on models of moderate size, including compact vision architectures and Transformer language models \cite{vaswani2017attention}. Its computational  overheads are of 5\%--55\% relative to vanilla SGD on standard CNN architectures.

\section{Preliminaries and Related Work}

\parhead{General Definitions.} 
We now briefly introduce the setting and notation; we refer the reader to standard texts~\cite{2014-martens, ly2017tutorial, amari2016information} for a complete introduction. 
We start from the standard setting in which we are given a dataset $\mathcal{D} = (\vec{x}_i, \vec{y}_i)_{i = 1}^N$, and wish to identify a $d$-dimensional model $\vw \in \R^d$ to minimize an empirical loss $L: \mathbb{R}^d \rightarrow \mathbb{R}$, defined as $L(\vec{\theta})=\frac{1}{N} \sum_{i=1}^{N} \ell\big(\vec{y}_i, f\left(\vec{x}_i; \vec{\theta}\right)\big)$. 
For a twice-differentiable loss $L$, the Hessian is the matrix of second-order derivatives of $L$ w.r.t. $\vw$, i.e. $\hess=\nabla^2_\vw L$. 
In the \emph{probabilistic} view, each input example $\vec{x}$ has some probability of being assigned a given label $\vec{y}$. 
Given input examples $\vx$ drawn from a distribution $Q_{\vx}$ and corresponding outputs drawn from a conditional distribution $Q_{\vy| \vx}$, the goal is to minimize the distance between the target joint distribution $Q_{\vx, \vy}= Q_{\vx} \,Q_{\vy| \vx}$, and a learned joint  distribution $P_{\vx, \vy}(\vw)$, where $\vw$ is the model.

\parhead{The Fisher Matrix.} 
Assuming the probabilistic view, it can be shown that the Fisher information matrix $\fisher$ of the model's joint distribution satisfies $\fisher=\mathrm{E}_{P_{\vx, \vy}}\left[- \nabla^2_\vw \log p_{\vx, \vy}(\vw)\right]\,$ where $p_{\vx, \vy}(\vw)$ is the density function. 
If the model's output conditional distribution matches the conditional distribution of the data, then the Fisher and Hessian matrices are in fact equivalent~\cite{ly2017tutorial}. 
Roughly, this means that, if the model has high accuracy, we can approximate the Hessian of $L$ at $\vw$ with the Fisher matrix. 

It is sometimes useful to consider an approximation to  the Fisher, where the distribution $P_{\vx, \vy}$ is replaced with the empirical training distribution $\widehat{Q}_{\vx, \vy}$, leading to the  \emph{empirical Fisher matrix}:  
{
	\begin{equation*}
	\begin{aligned}
	\widehat{\fisher} = & \mathrm{E}_{\widehat{Q}_{\vx}}\left[\mathrm{E}_{\widehat{Q}_{\vy | \vx}}\left[\nabla \log p_{\vy | \vx}(\vw) \nabla \log p_{\vy | \vx}(\vw)^{\top}\right]\right] 
	\stackrel{}{=}\frac{1}{N} \sum_{i=1}^{N} \underbrace{\nabla \ell\left(\vy_{i}, f\left(\vx_{i}; \vw\right)\right)}_{\nabla \ell_i} \nabla \ell\left(\vy_{i}, f\left(\vx_{i}; \theta\right)\right)^{\top}. 
	\end{aligned}
 	\end{equation*}
}

\parhead{The Woodbury-Sherman-Morrison Trick.} The fact that this approximation is a sum of rank-one matrices has the  benefit that it allows efficient exact computation of the Fisher inverse. 
Specifically, one can apply the classic Woodbury-Sherman-Morrison formula to compute the inverse recursively as
\begin{equation}\label{eq:woodfisher-inv}
\widehat{\fisher}^{-1} = \widehat{F}_{N}^{-1}=\widehat{F}_{N-1}^{-1}-\frac{\widehat{F}_{N-1}^{-1} \nabla \ell_{N} (\nabla \ell_{N}^{\top} \widehat{F}_{N-1}^{-1})}{N +\nabla \ell_{N}^{\top} \widehat{F}_{N - 1}^{-1} \nabla \ell_{N}}, 
\end{equation}
where the recursion is over samples, and the base step is $\widehat{F}_0^{-1} = \lambda^{-1} I_{d}$, with $\lambda$ being a small positive constant. 
This approach was independently introduced by Hassibi and Stork~\cite{1992-hassibi} for pruning, and Amari~\cite{1998-amari} for natural gradient.
Singh and Alistarh~\cite{2020-singh} showed that it can be scaled to DNNs by block-wise approximation, and that it provides better approximations of the loss than K-FAC-based~\cite{2019-wang}, or diagonal~\cite{2018-theis} approximations~\cite{2017-dong}, leading to more accurate pruning.  
The main shortcoming of directly applying (\ref{eq:woodfisher-inv}) is the prohibitive computational cost of $\Omega(d^2 m)$, even when reduced to $\Omega(dB m)$ by $B$-block-wise approximation.
Our method proposes a \emph{matrix-free} approach for exactly calculating IHVPs and querying individual elements of the inverse Hessian, of cost $O(dm)$ after $O(dm^2)$ precomputation. 
Our method is numerically equivalent to the direct computation above. 

\parhead{Diagonal Approximations.}
A common approximation, both for optimization, e.g.~\cite{krishnan2017neumann, kingma2014adam} but also for pruning~\cite{2018-theis}, is to assume that the Fisher matrix is \emph{diagonal}. 
In our notation, this method has setup cost $O(dm)$, and diagonal query cost $O(d)$. 
However, as evident from the experimental results of~\cite{2020-singh} this provides lower approximation quality relative to both block-wise methods or K-FAC~\cite{2019-wang}. 

\parhead{K-FAC Approximations.} 
This approach observes that the entries of the true Fisher corresponding to blocks ``between'' two layers, which can be written as the expectation of a Kronecker product between two matrices, can be approximated as the Kronecker product of the expectations of those two matrices (reversing the order between the product and the expectation). 
This approximation has been leveraged for both pruning and for optimization, and it allows efficient computation of the inverse~\cite{2016-ba, Osawa_2019,2019-zheng, 2019-wang,laurent2018an}; however, it is known to not always hold~\cite{2015-martens}. 
Another relative drawback is that the Kronecker factorization only occurs naturally for fully-connected layers; but there is work on extensions to other layer types~\cite{grosse2016kroneckerfactored, martens2018kroneckerfactored}, via additional approximations. We compare against K-FAC-based optimizers and pruners, and find that our method yields better results.

\parhead{Additional Approaches.} Our approach is similar in spirit to matrix-free methods~\cite{nocedal2006numerical, liu1989limited, martens_free}, but the algorithms we present are new.  \emph{Hessian-free} optimization~\cite{martens_free} also forgoes the explicit computation of Hessians in favor of computing an IHVP with a vector $\vv$. However, this estimation is performed by iteratively approximating the solution to the linear system $\mathbf{H} \vx = \vv$ for some given $\vx$ without ever explicitly forming $\mathbf{H}$.
One disadvantage of this method in practice is that it requires tuning and several very costly iterations to converge (for a single $\mathbf{v}$), as the underlying Hessian can be ill-conditioned. 
The L-OBS pruning method~\cite{2017-dong} approximates second-order information by defining independent layer-wise objectives, which allows the direct approximation of the layer-wise Hessians via a carefully-crafted block structure. In contrast, our approach allows for fully-global Hessian estimation and it also yields better pruning results at scale.

\emph{Full-matrix adaptive regularization} has similar goals, but in the context of adaptive optimizers~\cite{Duchi2010AdaptiveSM}. Agarwal et al.~\cite{agarwal2019efficient} proposed GGT, which allows the efficient computation of the \emph{inverse square root} of the low-rank matrix resulting from the sum of gradient outer products over a sliding window. 
At its core, this procedure requires  an eigen-decomposition (implemented via SVD) of an $m \times m$ matrix at every time step, which is reasonably efficient for small values of $m$.

Our dynamic algorithm solves a similar, but slightly simpler problem, as we only want to invert the matrix, without computing its square root. We do not perform any eigen-decompositions; instead, we carefully maintain intermediate information that allows an efficient explicit computation of the scalar coefficients in Equation~\ref{eq:dynamic-idea}. At the end of Section \ref{sec:dynamic}, we discuss how this approach is more efficient in practice and allows executing at larger window sizes (with small overhead), which leads to improved model accuracy as we show in our experiments.
Additionally, our dynamic algorithm has per-step cost $O( dm + m^3 ),$ relative to $O(dm^2 + m^3)$ reported by GGT~\citep{agarwal2019efficient}; this can result in lower overhead versus GGT, even when the SVD cost is negligible (e.g. if $m$ is small).

Yao et al.~\cite{yao2020adahessian} recently provided an alternative method for approximating the diagonal of the inverse Hessian, using Hutchinson's randomized algorithm for estimating the diagonal. To mitigate the variance, the authors introduce non-trivial smoothing heuristics. Their approach has theoretical per-iteration cost of at least $2\times$ versus SGD, as it requires a second backward pass over the network; and this cost is usually higher in practice. Experimentally, our algorithm often matches their accuracy, and, unlike AdaHessian, its cost depends exclusively on the model size, irrespective of the underlying structure. Thus, as we show in the experimental section, our average \emph{practical overhead} is less than 50\% for dense models, and less than 10\% for sparse ones.

\parhead{Approximation Quality.} Künstner et al.~\cite{2019-kunstner} performed an in-depth analysis of the empirical Fisher, making the point that, in theory, the approximation could become meaningless if all sample gradients are zero. Similarly to~\cite{2020-singh}, we did not find that this occurs in practice for deep neural networks, as sample gradients are never zero. The latter reference provides detailed comparisons of diagonal, K-FAC, and other approximations, and finds that the empirical Fisher can provide competitive  approximation quality for DNNs. Further, they demonstrate that better loss approximation implies better accuracy for neural network pruning. We therefore do not repeat their loss analysis, and mainly compare methods in terms of application performance.

\section{Algorithm Descriptions}

\subsection{The Static Algorithm}

As a warm-up, we first describe the IHVP algorithm given a static set of gradients. While our notation is customized for the empirical Fisher, all our techniques are applicable to any matrix that can be written as the sum of $m$ rank-one components.
Specifically, we are given $m$ vectors, which we assume to be the gradient vectors $(\nabla \ell_i)_{i = 1}^m$, and must compute quantities related to the inverse of the matrix resulting from the sum (more precisely, the average) of their outer products, which we denote by  $\widehat{F}_{m}^{-1}$. 
The main idea is to rewrite the recursive description of $\widehat{F}_{m}^{-1}$ such that, after some precomputation, the matrix-vector-product $\widehat{F}_{m}^{-1} \mathbf{x}$ can be computed efficiently for any vector $\mathbf{x}$. Using the Sherman-Morrison formula applied to the partial empirical Fisher matrix $\widehat{F}_{i}$ corresponding to the first $i$ gradients (and scaled by $1/m$), we obtain the following recursion: 
\begin{align}
    \label{eq:hinv-def}
    \widehat{F}_{i}^{-1} \vx = \widehat{F}_{i - 1}^{-1} \vx - 
    \frac{\widehat{F}_{i - 1}^{-1} \nabla \ell_i (\widehat{F}_{i-1}^{-1} \nabla \ell_i)^\top}{m + \nabla \ell_i^\top \widehat{F}_{i - 1}^{-1} \nabla \ell_i} \vx, 
    \textnormal{ with } \widehat{F}_0^{-1} \vx = \lambda^{-1} I_d \vx = \lambda^{-1} \vx \textnormal{ and } \lambda > 0.
\end{align}
For simplicity, let us set $\mathbf{v_i} = \widehat{F}_{i - 1}^{-1} \nabla \ell_i$ and unroll the Sherman-Morrison recursion, which gives:
\begin{align}
    \widehat{F}_i^{-1} \vx 
    = \widehat{F}_{i - 1}^{-1} \vx - \mathbf{v_i} \frac{\mathbf{v_i}^\top \vx}{m + \nabla \ell_i^\top \mathbf{v_i}} 
    = \lambda^{-1} \vx - \sum^i_{j = 1} \mathbf{v_j} \frac{\mathbf{v_j}^\top \vx}{m + \nabla \ell_j^\top \mathbf{v_j}}. \label{eq:unrolled-mul}
\end{align}
\parhead{IHVP Computation.} Assuming that we have already computed all the vectors $\mathbf{v_j}$, the above expression can be calculated in $O(dm)$ time without any intermediate $d \times d$ matrices: first compute the scalar fractions in~(\ref{eq:unrolled-mul}), and then evaluate the resulting linear combination of $\vx$ and $\mathbf{v_j}$. 
Since $\mathbf{v_i} = \widehat{F}_{i - 1}^{-1} \nabla \ell_i$, it can be computed in exactly the same way given all $\mathbf{v_j}$ for $j < i$. Thus, all $m$ vectors $\mathbf{v_i}$ of dimension $d$ can be precomputed in increasing order of $i$ using $O(dm^2)$ total time. As an additional optimization (to cut the required memory in half), we also precompute all $q_j = m + \nabla \ell_j^\top \mathbf{v_j}$.

\parhead{Querying the Inverse.} 
To extract individual elements $\lbrack \widehat{F}_{m}^{-1}\rbrack_{ij}$ of the inverse Hessian approximation, we can write $\mathbf{e_i}^\top \widehat{F}_{m}^{-1} \mathbf{e_j}$ in the form of (\ref{eq:unrolled-mul}) where $\mathbf{e_i}$ and $\mathbf{e_j}$ are indicator vectors
\begin{equation}
    \mathbf{e_i}^\top \widehat{F}_m^{-1} \mathbf{e_j} = \mathbf{e_i}^\top\lambda^{-1} \mathbf{e_j} - \sum^m_{k = 1} \mathbf{e_i}^\top \mathbf{v_k} \frac{\mathbf{v_k}^\top \mathbf{e_j}}{q_k}. \label{eq:unrolled}
\end{equation}
As $\mathbf{e_i}^\top \mathbf{v_k}$ and $\mathbf{v_k}^\top \mathbf{e_j}$ can both be realized as constant time indexing operations, the above turns into a sum over $m$ scalars. Hence, our method admits $O(m)$ access to any element of $\widehat{F}_{m}^{-1}$ using the same precomputed $\mathbf{v_k}$ and $q_k$ as for the efficient calculation of $\widehat{F}_{m}^{-1} \vx$.

\parhead{Additional Optimizations.} 
The algorithm admits a fast vectorized implementation, and several optimizations, which we describe in the Appendix. For example, we perform several memory-saving optimizations, as well as explicit page swapping between CPU and GPU memory to mitigate the gradient transfer costs. 
Furthermore, the static algorithm can be applied independently to each block of a block-wise approximation: for block size $B$, the computation and memory costs per block are reduced by $B/d$, but since there are now $d/B$ blocks, the overall costs will stay the same irrespective of $B$. Thus, as long as $m < B$, our method is $B/m$ times faster (and less memory intense) than the direct implementation of the Woodbury inverse.

\subsection{The Dynamic Algorithm}
\label{sec:dynamic}

We now describe the \emph{dynamic} algorithm, which assumes that gradients arrive in an online fashion and must be integrated into the Fisher estimate. 
We first present the algorithm itself, i.e. the setup / update / IHVP computations, and perform a complexity analysis. Next, we show how the algorithm can be derived and finally we conclude with notes on an efficient practical implementation.

\parhead{High-Level View.} The main idea of the dynamic algorithm is to write the IHVP as
\begin{equation}
    \label{eq:dynamic-idea} 
    \widehat{F}_{m}^{-1} \vx = \lambda^{-1} \vx - \sum^m_{j=1} c^m_j \nabla \ell_{j},
\end{equation}
where the scalar coefficients $c^m_j$ can be computed efficiently from just the scalar products $\nabla \ell_i^\top \nabla \ell_j$ and $\nabla \ell_i^\top \vx$. Then, any gradient $\nabla \ell_{i}$ can be replaced by updating just $O(m)$ of the stored scalar product values.
In the following, we use $\mathbf{G} = [\nabla \ell_1^\top; \nabla \ell_2^\top; \dots; \nabla \ell_m^\top]$ to denote the row-wise $m \times d$ matrix of the $m$ gradients for which we wish to compute the inverse empirical Fisher. Further, for each $i$ from 1 to $m$, let $\mathbf{b^i}$ be a vector with $m$ components $(b^i_1, b^i_2, \ldots, b^i_m)$, such that 
\begin{equation}
    \widehat{F}_{i - 1}^{-1} \nabla \ell_i = \lambda^{-1} \nabla \ell_i - \sum_{j = 1}^{i-1} a^{i - 1}_j \nabla \ell_j = \sum_{j = 1}^m b^i_j \nabla \ell_j,
\end{equation}
i.e. containing the scalar coefficients of $\widehat{F}_{i - 1}^{-1} \nabla \ell_i$. This means that $b^i_j = -a^{i - 1}_{j}$ when $j < i$, $b^i_i = \lambda^{-1}$, and $b^i_{j} = 0$ otherwise.

\parhead{Initial Precomputation \& Update.}
The dynamic algorithm maintains three $m \times m$ matrices: $\mathbf{G}\mathbf{G}^\top$, $\mathbf{D}$ and $\mathbf{B}$. The first is the symmetric gradient scalar product matrix $[\mathbf{G}\mathbf{G}^\top]_{ij} = \nabla \ell_i^\top \nabla \ell_j$. The second is an upper triangular matrix and stores the precomputed values $[\mathbf{D}]_{ij} = \nabla \ell_i^\top \widehat{F}_{i - 1}^{-1} \nabla \ell_j$ for $i \leq j$. The third is the row-wise matrix of the coefficient vectors $[\mathbf{B}]_{ij} = b^i_j$, which makes it lower triangular with a diagonal of $\lambda^{-1}$. We now discuss how to compute those matrices.

The initial setup of the dynamic algorithm begins by evaluating $\mathbf{G}\mathbf{G}^\top$ in a straightforward fashion. Next, $\mathbf{D}$ can be computed for $i \leq j$ according to the following recursion:
\begin{align}
    [\mathbf{D}]_{ij} &= (\nabla \ell_i^\top  \widehat{F}_{i - 1}^{-1} \nabla \ell_j) \textnormal{ and } (\nabla \ell_j^\top \widehat{F}_{0}^{-1} \nabla \ell_k) = \lambda^{-1} [\mathbf{G}\mathbf{G}^\top]_{jk} \\
    (\nabla \ell_j^\top  \widehat{F}_{i}^{-1} \nabla \ell_k)  &= (\nabla \ell_j^\top \widehat{F}_{i - 1}^{-1} \nabla \ell_k) - \frac{(\nabla \ell_j^\top \widehat{F}_{i - 1}^{-1} \nabla \ell_i)(\nabla \ell_i^\top \widehat{F}_{i - 1}^{-1} \nabla \ell_k)}{m + (\nabla \ell_i^\top \widehat{F}_{i - 1}^{-1} \nabla \ell_i)}. 
\end{align}
Given $\mathbf{D}$, we can then conclude the precomputation by calculating $\mathbf{B}$ for $i \geq j$ recursively as:
\begin{align}
    [\mathbf{B}]_{ii} &= \lambda^{-1} \,\,\textnormal{ and }\, \, [\mathbf{B}]_{ij} = -\sum_{k = j}^{i - 1} \frac{[\mathbf{D}]_{ki}}{m + [\mathbf{D}]_{kk}} [\mathbf{B}]_{kj} \label{eq:B-formula}.
\end{align}
After the initial setup, gradient $\nabla \ell_k$ can be replaced with gradient $\nabla \ell'_k$ by first updating row $k$ of $\mathbf{G}$ and then replacing row and column $k$ in $\mathbf{G}\mathbf{G}^\top$ with $\mathbf{G} \nabla \ell'_k$. Afterwards, the recomputation of columns $j \geq k$ of $\mathbf{D}$ and rows $i \geq k$ of $\mathbf{B}$ completes the update.

\parhead{Multiplication.}
Once $\mathbf{G}\mathbf{G}^\top$, $\mathbf{D}$ and $\mathbf{B}$ have been precomputed, one can perform efficient IHVPs of the form $\widehat{F}_{m}^{-1} \mathbf{x}$ with arbitrary vectors $\vx$. This is done by first evaluating $\mathbf{p} = \mathbf{G} \mathbf{x}$ and then computing all $m$ values $q_i$ by the following recursion:
\begin{align}
    q_i &= \frac{(\nabla \ell_i^\top  \widehat{F}_{i - 1}^{-1} \mathbf{x})}{m + [\mathbf{D}]_{ii}} \textnormal{ and }  (\nabla \ell_j^\top \widehat{F}_{0}^{-1} \mathbf{x}) = \lambda^{-1} p_j\\
    (\nabla \ell_j^\top  \widehat{F}_{i}^{-1} \mathbf{x})  &= (\nabla \ell_j^\top \widehat{F}_{i - 1}^{-1} \mathbf{x}) - \frac{[\mathbf{D}]_{ij}}{m + [\mathbf{D}]_{ii}}(\nabla \ell_i^\top \widehat{F}_{i - 1}^{-1} \mathbf{x}).
\end{align}
Eventually, the final result of the IHVP is obtained by:
\begin{equation}
    \widehat{F}_{m}^{-1} \mathbf{x} = \lambda^{-1}\mathbf{x} - \sum_{j = 1}^m (\sum_{k=j}^{m} q_k [\mathbf{B}]_{kj}) \nabla \ell_j. \label{eq:ihvp-formula}
\end{equation}

\parhead{Complexity Analysis.} The dynamic algorithm stores $m$ gradients of dimension $d$ as well as three $m \times m$ matrices, and thus has an overall memory complexity of $O(dm + m^2)$.

Next, we analyze the time complexity of all important operations. Initially, $\mathbf{G} \mathbf{G}^\top$ must be computed once, which takes $O(dm^2)$ time. 
Then, the recursion of $\mathbf{D}$ has three indices with $1 \leq i, j, k \leq m$ and each step takes constant time. Thus, it can be computed in $O(m^3)$ time with dynamic programming. Further, since values for index $i$ depend only on values for index $i - 1$, the dynamic programming can be implemented in $O(m^2)$ space. $\mathbf{B}$ has two indices $1 \leq i, j \leq m$ and every recursion takes $O(m)$ time, meaning that it can also be fully computed in $O(m^3)$ through dynamic programming. Hence, the overall initial setup cost is $O(dm^2 + m^3)$.

To replace one gradient with $\nabla \ell'$, we have to compute $\mathbf{G} \nabla \ell'$ as well as (partially) recalculate $\mathbf{D}$ and $\mathbf{B}$, which takes at worst $O(dm + m^3)$ time. An IHVP requires two matrix-vector products involving $\mathbf{G}$ and a recursion with two indices and therefore has a complexity of $O(dm + m^2)$.

\parhead{Algorithmic Derivation.} 
The dynamic algorithm can be derived directly from Theorem \ref{thm:dynamic}, which we state here in a simplified form (using the definitions of $\mathbf{B}$ and $\mathbf{D}$), and prove in the Appendix.

\begin{theorem}
    \label{thm:dynamic}
    Let $\widehat{F}_{i} = \lambda I + 1/m \cdot \sum_{j = 1}^i \nabla \ell_j \nabla \ell_j^\top$, then $\widehat{F}_{i}^{-1} \vx$ can be calculated as:
    \begin{equation}
        \label{eq:mul-coefs}
        \widehat{F}_{i}^{-1} \vx = \lambda^{-1} \vx - \sum^i_{j=1} c^i_j \nabla \ell_{j}  \,\,\textnormal{ and }\, \,  c^i_j = \sum_{k=j}^{i} \frac{(\nabla \ell_k^\top \widehat{F}_{k - 1}^{-1} \vx)}{m + [\mathbf{D}]_{kk}} [\mathbf{B}]_{kj}.
    \end{equation}
\end{theorem}

Equation (\ref{eq:mul-coefs}) with $i = m$ is exactly equal to the IHVP computation (\ref{eq:ihvp-formula}) as the fraction in the innermost sum corresponds to $q_k$. Similarly, index shifting $i = i - 1$ and setting $x = \nabla \ell_i$, thus turning $c^i_j$ into $-[\mathbf{B}]_{ij}$ as well as the fraction's numerator to $[\mathbf{D}]_{ki}$, recovers the precomputation formula of $\mathbf{B}$ given by (\ref{eq:B-formula}) for $i > j$. The formulas for $\mathbf{D}$ and $q_i$ follow directly from an expansion of the Woodbury formula followed by an appropriate recursive evaluation that avoids any matrix / vector operations except in the base case (we indicate the corresponding recursive calls by brackets).

\parhead{Efficient Implementation.} Directly implementing the discussed recursive formulas in a modern machine learning framework would result in very slow code. Fortunately, it is possible to implement all computations required for the dynamic algorithm very efficiently on a GPU. We describe how to do this in the Supplementary Material and provide a full implementation \cite{Code}. Specifically, we provide complete sample code that is able to perform the calculation of $\mathbf{D}$ and $\mathbf{B}$, i.e. the $O(m^3)$ component of the overall update cost, in $< 10$ milliseconds (on an NVIDIA RTX 2080 Ti) for values of $m$ as high as $1024$ (and the code can still be further optimized). For reference, this is $> 10\times$ faster than the highly-optimized $m \times m$ SVD  computation done at every step by GGT \cite{agarwal2019efficient}. We emphasize that this $O(m^3)$ computation being very fast in practice is crucial to reach low overheads, especially when dealing with a  models where the $O(dm)$ matrix-vector products are not the bottleneck.

\section{Experimental Validation}

\subsection{Application 1: Pruning DNNs using the Static Algorithm}
\parhead{Background.}
Given a trained model $\vw^{\star}$, the goal of pruning is to find the weight $\w_i$, or the set of weights, whose setting to $0$ would lead to a minimal increase in the training loss. 
Under a local quadratic approximation of the loss function, the OBD framework~\cite{1990-lecun} shows that the ``optimal'' weight to be removed is the one with the lowest value of the saliency metric $\rho(\theta_i)  = \frac{\theta_i^2}{2 \, \lbrack\hessinv\rbrack_{ii}}$ and proposed to estimate $\lbrack\hessinv\rbrack_{ii}$ by diagonal approximation. 
The OBS framework~\cite{1992-hassibi} observed that the \emph{remaining} weights $\w$ should also be updated via the optimal perturbation
$\delw  = -\w_i \hessinv \e_i / \lbrack\hessinv\rbrack_{ii},$
where $\e_i$ is the $i$th basis vector. Our algorithm efficiently supports both these operations.

Wang et al.~\cite{2019-wang} raised the valid point that applying the OBS update above to multiple weights being removed at once may be incorrect, since it ignores possible correlations between those weights.  
We considered this point in detail, comparing results between OBD pruning~\cite{1990-lecun}, the OBS update~\cite{1992-hassibi}, and an augmented version of OBS which disentangles the correlations by solving a corresponding linear system~\citep{2020-singh}. 
The results, presented in the Appendix, suggest that the OBS update is quite beneficial even in its approximate form, and that the effect of correlation is small for unstructured pruning. In fact, we find that the OBS pruning mask in each step is usually very similar to the one obtained by simple magnitude pruning, suggesting that the final accuracy improvements are primarily due to the updates of the remaining weights, facilitating better recovery.

\parhead{Experimental Setup.} We prune CNNs (ResNet-50~\cite{He_2016} and MobileNet-V1~\cite{howard2017mobilenets}) on the ImageNet dataset~\cite{russakovsky2015imagenet}.
These models are standard in the pruning literature~\cite{gale2019state}, and therefore several strong baselines exist.
Timing experiments are run on a machine with NVIDIA RTX 2080 Ti GPUs, a 48-core Intel CPU, and 512 GB of RAM. 
Following~\citep{2020-singh}, we used batched gradients (of size 16) as single samples inside the Fisher approximation. This does not alter results, but reduces variance.

We compare against Global Magnitude Pruning (GMP)~\cite{1994-hagiwara}, Layer-wise Optimal Brain Surgeon (L-OBS)~\cite{2017-dong},  Soft Threshold Reparametrization (STR)~\cite{2020-kusupati}, and  WoodFisher (WF)~\cite{2020-singh}.
The latter two methods are state-of-the-art for gradual pruning. WF is numerically equivalent to our method at the same parameter settings, but has significantly higher  computational and storage cost.
Relative to it, we therefore focus on executing at higher parameter values. We compare against their public implementation.
The Appendix contains full hyper-parameters, ablations with respect to block size and number of gradients, and a comparison with K-FAC pruning in a simpler setting~\cite{2019-zheng}.

We first perform a one-shot comparison to evaluate the ``raw'' per step pruning performance of M-FAC relative to other methods. Next, we evaluate how simply increasing M-FAC parameter values (but keeping everything else exactly the same as in WF's experiments \cite{2020-singh} which follow \citep{2017-zhu}) improves over state-of-the-art gradual pruning results by WF, GMP and STR. Finally, we demonstrate how M-FAC's high per-step efficiency can be utilized to craft practical pruning schedules with little computational overhead relative to GMP and only moderate extra memory consumption, which at the same time yield significantly better results than WF's state-of-the-art numbers.

\begin{figure}[t]
    \centering
    \begin{subfigure}[b]{0.42\textwidth}
        \centering
        \includegraphics[width=\textwidth]{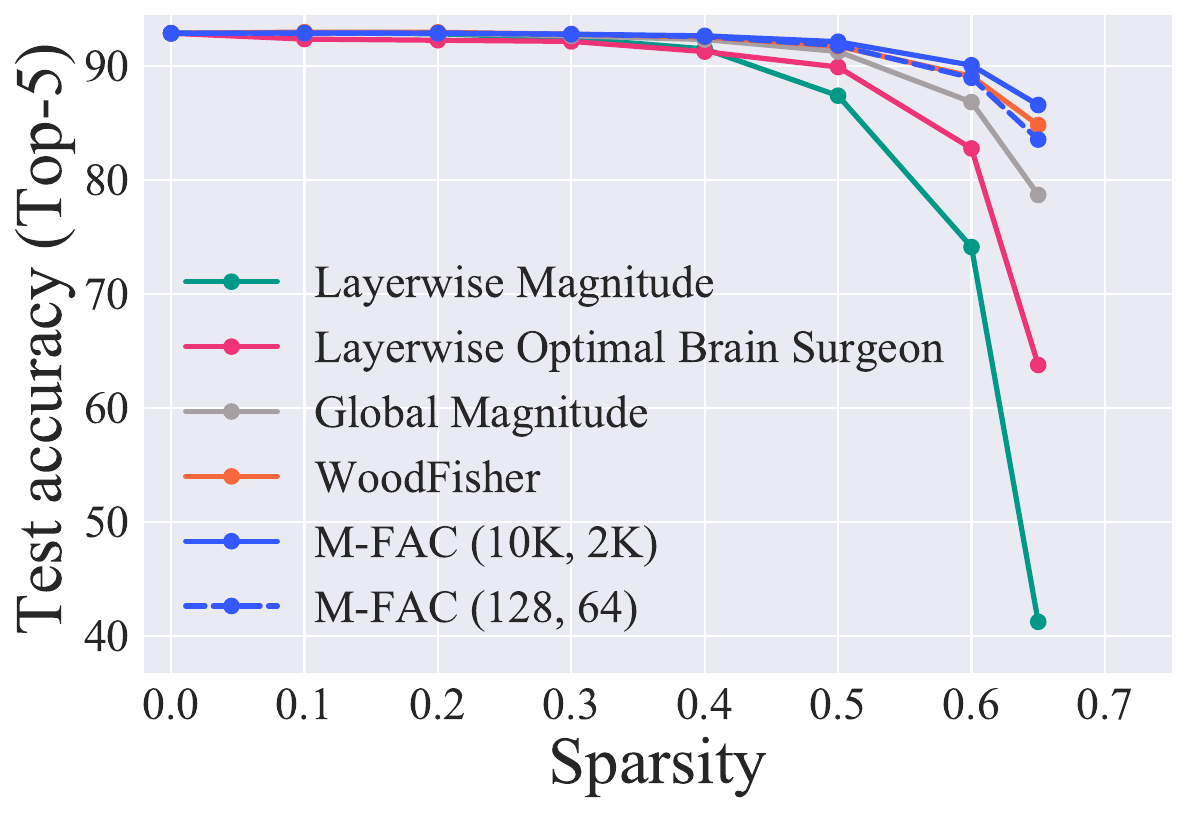}
        \caption{One-shot pruning for ResNet50.}
        \label{fig:oneshot-rn50}
    \end{subfigure}
    \hfill
    \begin{subfigure}[b]{0.42\textwidth}
        \centering
        \includegraphics[width=\textwidth]{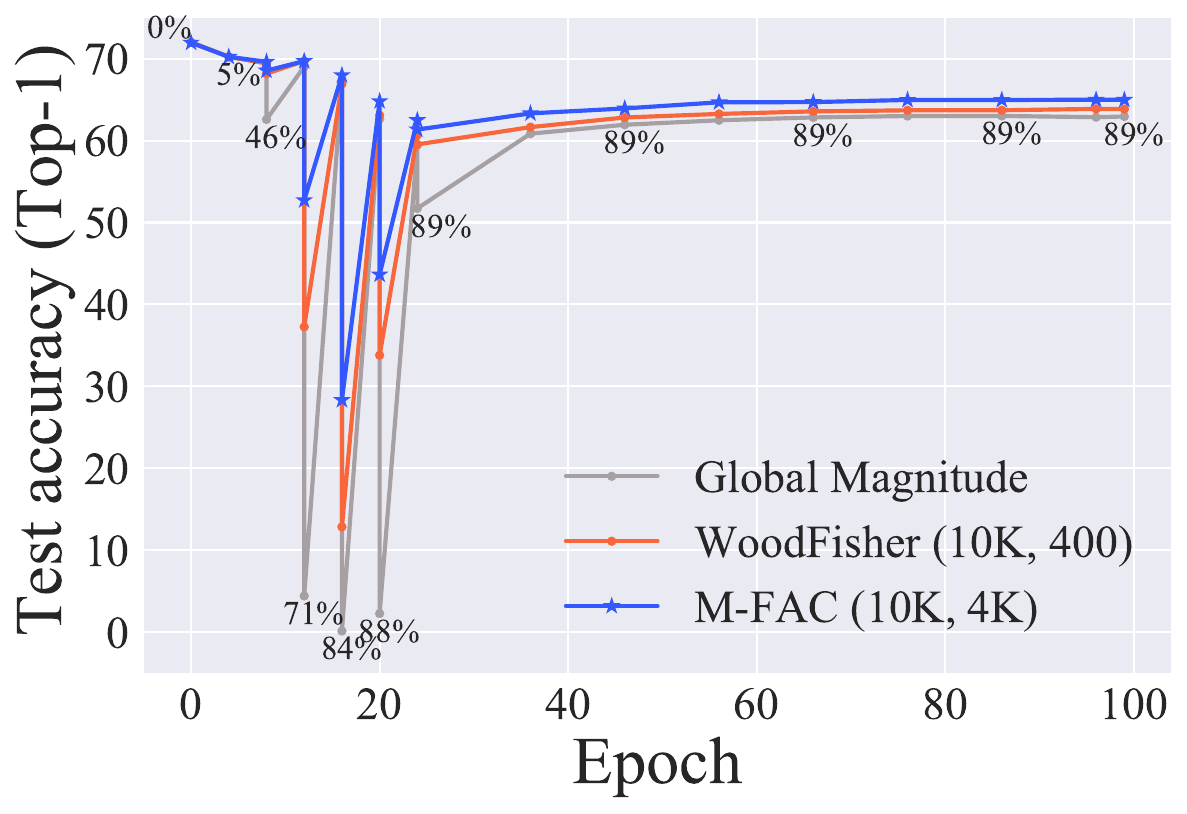}
        \caption{Gradual pruning for MobileNetV1-STR.}
        \label{fig:gradual-mobilenet}
    \end{subfigure}
    \vskip\baselineskip
    \begin{subfigure}[b]{0.42\textwidth}
        \centering
        \includegraphics[width=\textwidth]{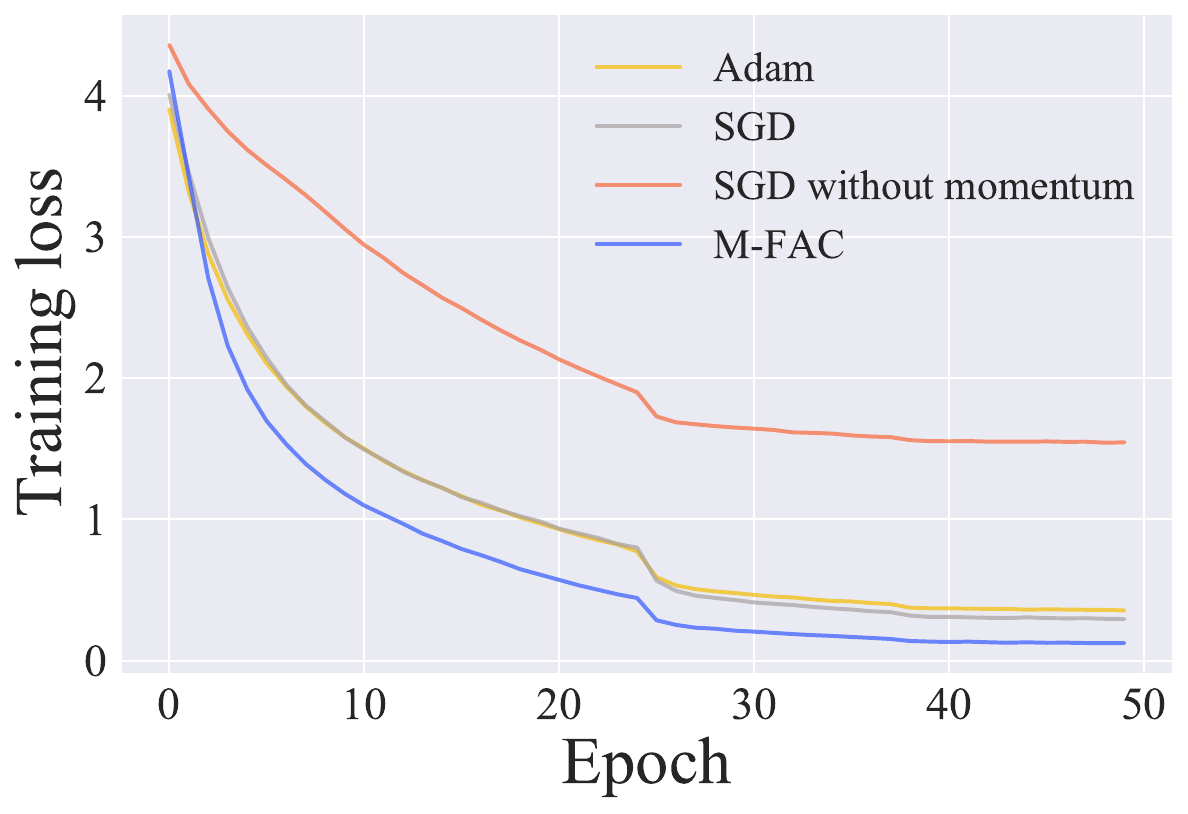}
        \caption{Training loss for WRN-40-2 / CIFAR-100.}
        \label{fig:opt-loss}
    \end{subfigure}
    \hfill
    \begin{subfigure}[b]{0.42\textwidth}
        \centering
        \includegraphics[width=\textwidth]{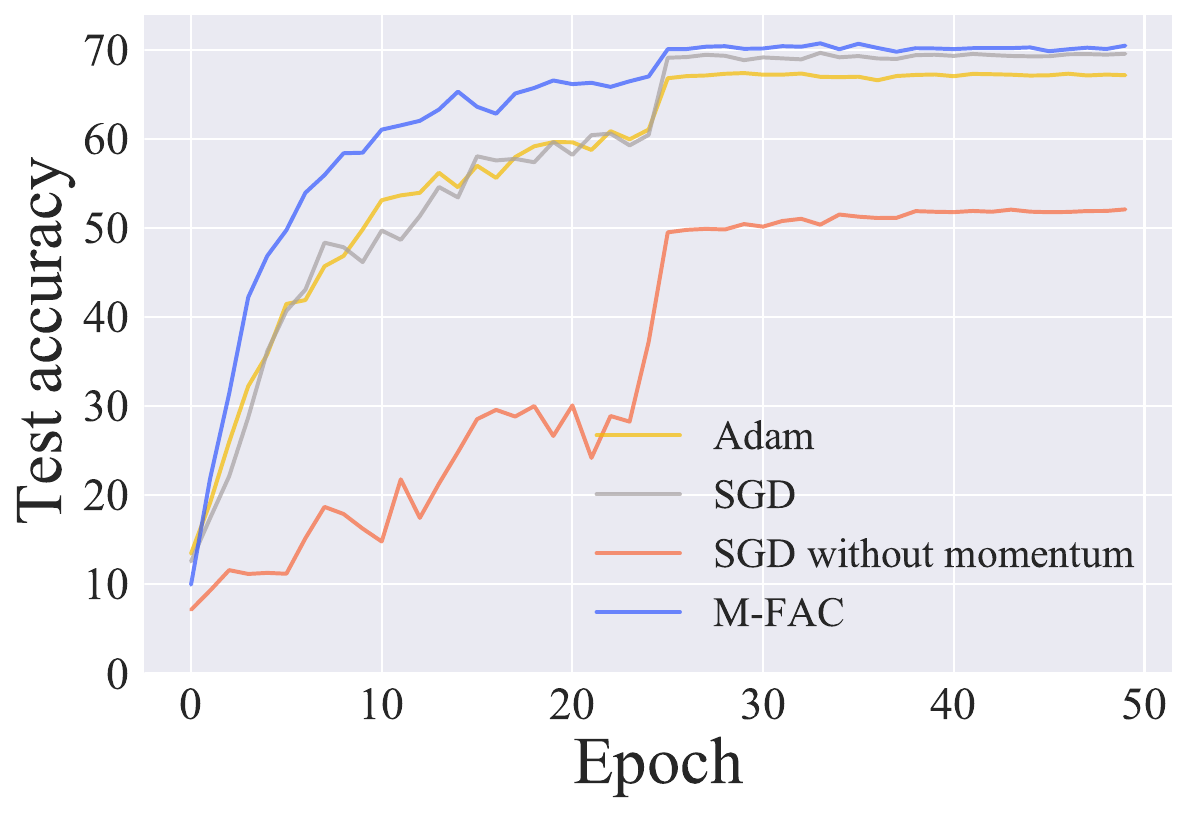}
        \caption{Test accuracy for  WRN-40-2 / CIFAR-100.}
        \label{fig:opt-acc}
    \end{subfigure}
    \caption{\textbf{(a)} Accuracy vs. sparsity for one-shot pruning on ResNet50 / ImageNet. \textbf{(b)} Accuracy during gradual pruning for the MobileNetV1-STR / ImageNet experiment. \textbf{(c)} and \textbf{(d)} Comparison of different optimizers on a short WRN-40-2 / CIFAR-100 schedule.}
\end{figure}

\parhead{Oneshot Pruning.} Figure~\ref{fig:oneshot-rn50} shows the Top-5 accuracy of one-shot ResNet50 pruned models for L-OBS, GMP, WoodFisher, and M-FAC, where our method is executed with a parameter setting that is infeasible for WoodFisher due to the required computational and storage costs.
(We use Top-5 accuracy to compare with~\cite{2017-dong}.)
We note the improved accuracy of global methods relative to layer-wise, and that the Fisher approximation yields consistently better results relative to global magnitude.
This suggests that a better approximation of the Fisher inverse also yields better pruning results, and is in line with the loss approximation study of~\citep{2020-singh}.

Generally, we found that estimation using more gradients always improves results, until saturation.
Interestingly, increasing the block size does not necessarily improve results, for some models, smaller block sizes sometimes yield better results for a fixed number of gradients.

Figure~\ref{fig:gradual-mobilenet} examines the effect of the improvement in one-shot pruning accuracy at each step of pruning, on the final result, when pruning MobileNetV1-STR gradually to $89\%$ sparsity. WoodFisher uses block size $10K$, and $m = 400$ gradients, while M-FAC uses the same block size but $m = 4000$ gradients. Executing WoodFisher with the same parameters would be extremely slow. Note the gap in one-shot accuracy following each pruning step: this translates in the final accuracy gap of more than 1\% even after the extensive fine-tuning phase. 

\parhead{Gradual Pruning Comparisons.} Table~\ref{tab:gradual-sota} presents gradual pruning results for MobileNetV1-STR/ImageNet at 89\% sparsity and ResNet50/ImageNet at 95\% sparsity, relative to other state-of-the-art methods. M-FAC outperforms previous methods in the case of MobileNetV1-STR by more than 1.5\% Top-1 accuracy. Further, despite using $10 \times$ more gradients to estimate the empirical Fisher, the per-step pruning cost of M-FAC is $> 6 \times$ lower than WoodFisher's as can be seen in Table \ref{tab:pruning-speed} (for the same parameters settings M-FAC is $> 100\times$ faster). The ResNet50 results show smaller gains.

\begin{minipage}[c]{0.55\textwidth}
	\centering
	\captionof{table}{\footnotesize{Gradual pruning results (top-1 accuracy) for \textbf{MobileNetV1-STR / ImageNet}  at 89\% sparsity and \textbf{ResNet50 / ImageNet} at 95\% sparsity.}}
	\scalebox{0.85}{
		\begin{tabular}{@{}lccccc@{}}
			\toprule
			Method &  MBv1 Top-1 (\%)   & RN50 Top-1 (\%)  \\
			\midrule
            Dense baseline & 72.00  & 77.01   \\
            \midrule
			STR \citep{2020-kusupati} & 62.10 & 70.40 \\
			Global Magnitude &  62.94 & 71.72  \\
			{{WoodFisher}}~\cite{2020-singh} & {{63.87}} & 72.12  \\
			\textbf{{M-FAC}}  &  \textbf{{65.01}} & \textbf{72.32} \\
 			\midrule
		\end{tabular}
		\label{tab:gradual-sota}
	}
\end{minipage}~
\begin{minipage}[c]{0.4\textwidth}
	\centering
	\captionof{table}{\small{Minutes per pruning step compared with WoodFisher (WF). OOM denotes out of memory on a system with 512GB RAM.}}
	\scalebox{0.85}{
        \begin{tabular}{l|c|c|r|r}
            \toprule
            Model & $B$ & $m$ & WF & \textbf{M-FAC} \\
            \midrule
            MBv1-STR & 10k & 400 & 60m & \textbf{0.5m} \\
            MBv1-STR & 10k & 4k & OOM & \textbf{8.7m} \\
            MBv1-STR & all & 1k & OOM & \textbf{1.2m} \\
            ResNet50 & 2k & 400 & 35m & \textbf{2.5m} \\
            ResNet50 & 10k & 1k & OOM & \textbf{14.0m} \\
            ResNet50 & all & 512 & OOM & \textbf{4.7m} \\
            \bottomrule
        \end{tabular}
        \label{tab:pruning-speed}
    }
\end{minipage}

\parhead{Practical Pruning.}
Due to the computational cost of a WF step, the previous schedules used only a small number of pruning steps. However, M-FAC makes it possible to prune much more frequently. 
In this case, the cost of extracting and storing several thousand gradients may still be a bottleneck. We find that M-FAC already performs well with low parameter settings, especially when coupled with recomputations (i.e. resampling gradients and rebuilding the M-FAC approximation after partial pruning steps). We now perform MobileNet and ResNet50 gradual pruning runs using M-FAC with block size $B = 128$, 64 gradients and 16 recomputations per step. A single M-FAC pass takes less than a few seconds at these settings.  Details about the pruning schedule can be found in the Appendix. This is a practical setup, which can scale to larger models. At the same time, Table \ref{tab:practical-pruning} shows that leads to significantly improved pruning results, sometimes even by several percent accuracy, compared to WF and the M-FAC results in Table \ref{tab:gradual-sota}.

\begin{table}[h]
    \centering
    {\small
    \begin{tabular}{lcccccc}
        \toprule
        & MBv1-0\% & MBv1-75\% & MBv1-89\% & RN50-0\% & RN50-95\% & RN50-98\% \\
        \midrule
        WoodFisher & 72.0 & 70.1 & 63.9 & 77.0 & 72.1 & 65.6 \\
        \textbf{M-FAC} & 72.0 & \textbf{71.0} & \textbf{67.3} & 77.0 & \textbf{72.6} & \textbf{67.5} \\
        \bottomrule
    \end{tabular}
    }
    \caption{M-FAC performance with optimized practical settings.}
    \label{tab:practical-pruning}
\end{table}

\subsection{Application 2: Optimization using the Dynamic Algorithm}
\parhead{Matrix-Free Preconditioned SGD.}
The dynamic algorithm can provide an efficient implementation for the following variant of SGD. 
Let $t$ be the current iteration index, and $\Delta < t$ be the length of the sliding window during which the estimation of the inverse Hessian is performed. Consider the iteration
$    \vw_{t + 1} = \vw_{t} - \eta_t \widehat{F}^{-1}_{[t - \Delta, t]} \nabla \ell_t,$ where $\eta_t$ is the learning rate. 
Let $\nabla \ell_{t-\tau + 1}, \ldots, \nabla \ell_{t}$ be the gradients obtained at steps $(t-\tau + 1), \ldots, {t}$, respectively. 
Then, we define the preconditioning matrix as  
$\widehat{F}_{[t - \Delta, t]} = \lambda^{-1} I_d + 1/\Delta \cdot \sum_{\tau = 0}^{\Delta - 1} \nabla \ell_{t - \tau} \cdot \nabla \ell_{t - \tau}^{\top},$ that is, an empirical Fisher matrix generated with respect to the sliding window of $\Delta$ gradients. 
At each step, our dynamic algorithm replaces $\nabla \ell_{t - \Delta + 1}$ with $\nabla \ell_t$ in the inverse estimate, and then computes $\widehat{F}^{-1}_{[t - \Delta, t]} \nabla \ell_t$. This corresponds to full-matrix natural gradient SGD \cite{1998-amari} under the extra assumption that gradients do not change too quickly during training. This assumption has been validated in the K-FAC literature \cite{2016-ba} and we also provide additional experimental justification in the Appendix. Finally, to better study the effect of the preconditionier, we do \textit{not} apply momentum to M-FAC in any of the experiments below.

\parhead{Initial Test.} We begin by verifying that the M-FAC optimizer actually behaves as one would expect from a second-order method, i.e. learning faster than standard first-order techniques, especially in the early epochs. For that purpose, we run M-FAC, SGD with and without momentum and Adam on WideResNet 40-2 / CIFAR10 with a compressed 50 epoch schedule. The corresponding Figures \ref{fig:opt-loss} and \ref{fig:opt-acc} indeed showcase the expected behavior. Interestingly, they also show that SGD without momentum performs poorly in this setup, which further highlights the major impact of the M-FAC preconditioner, which seems to be able to  make up for the lack of momentum.

\parhead{Second Order Comparison.} Next, we work with the common ResNet20 and ResNet32 models~\citep{He_2016} and compare against standard optimizers such as SGD with momentum, Adam~\citep{kingma2014adam} and AdamW~\citep{2019-loshchilov}, but also against approximate second-order methods such as K-FAC~\cite{martens2018kroneckerfactored}, GGT~\citep{agarwal2019efficient}, and AdaHessian~\citep{yao2020adahessian}.
For the former two, we use the implementations contributed to TensorFlow~\citep{abadi2016tensorflow} by the authors, while for the latter we use the authors' PyTorch implementation. For fairness, we always follow the settings recommended by the authors, although this makes overheads harder to compare: we used specific learning rates, and grid-searched the weight-decay values for each method in turn.

Since implementations have slightly different framework requirements, our main performance metric is \emph{overhead over SGD}, measured in each algorithm's environment, on an NVIDIA RTX 3090 GPU for PyTorch and a Titan RTX for TensorFlow. For all second-order methods, we compute the preconditioner at each step---reducing the frequency reduces overheads proportionally for each method, but introduces an additional hyper-parameter and generally results in lower final accuracy. We provide additional information in the Supplementary.

\begin{table}
    \begin{minipage}[c]{0.55\textwidth}
        \scalebox{0.8}{
            \begin{tabular}{@{}lccccc@{}}
                \toprule
            	\multirow{1}{*}{} & \multicolumn{2}{c}{ResNet20} & \multicolumn{2}{c}{ResNet32} \\
            	\cmidrule(l{3pt}r{3pt}){2-3}
            	\cmidrule(l{3pt}r{3pt}){4-5}
            	\multirow{1}{*}{Method} & Acc. & Overhead$^*$ & Acc. & Overhead$^*$ \\
            	\midrule
            	SGD & 91.78 & $1.00 \times$ & 92.80 & $1.00 \times$ \\
            	Adam & 89.67 & $1.05 \times$ & 90.56 & $1.10 \times$ \\
            	AdamW & 91.78 & $1.05 \times$ & 92.58 & $1.10 \times$ \\
            	\midrule
            	K-FAC & 91.65 & $1.21\times$ ($2.35\times$) & 90.09 & $1.50\times$ ($3.05\times$) \\
            	GGT & 88.38 & $1.05\times$ ($8.35\times$) & 89.14 & $1.05\times$ ($8.88\times$) \\
            	AdaHessian & 92.17 & $3.50 \times$ & \textbf{92.81} & $3.30 \times$ \\
            	M-FAC & \textbf{92.34} & $1.55 \times$ & 92.65 &  $1.50 \times$\\
             	\bottomrule
            \end{tabular}
        }
    \end{minipage}
    \begin{minipage}[c]{0.425\textwidth}
        \scalebox{0.8}{
            \begin{tabular}{@{}lccc@{}}
            	\toprule
            	Model&  SGD & Adam & M-FAC (Overh.)  \\
            	\midrule
            	WRN 22-2 & \textbf{69.93} & 66.90 & 69.76 ($1.40\times$)  \\
            	WRN 40-2 & 71.75 & 70.14 & \textbf{72.42} ($1.35\times$) \\
            	WRN 22-4 & 73.13 & 72.52 & \textbf{74.06} ($1.43\times$)\\
            	MBv1 Dense & 68.06 & 67.92 & \textbf{68.96} ($1.25\times$) \\
            	\midrule
            	MBv1 Sparse & 64.11 & -- & \textbf{64.78} ($1.03\times$) \\
            	RN50 Sparse & 74.78 & -- & \textbf{75.10} ($1.05\times$) \\
                \bottomrule
            \end{tabular}
        }
    \end{minipage}
    \caption{\textbf{Left}: Comparison of M-FAC against first- and second-order optimizers with individual tuning and weight decay. \textbf{Right}: Additional experiments with no tuning and weight decay. Sparse results are from finetuning after the last step of gradual pruning to $\approx 90\%$ sparsity. $^*$ For K-FAC / GGT we indicate in parentheses the overhead in a comparable setting to M-FAC with the same batchsize / the same number of gradients.}
    \label{tab:optimizer-results}
\end{table}

Our first results are presented in Table 4 (left), where we examine the best Top-1 test accuracy obtained by each method, as well as its overhead relative to SGD. For K-FAC, the best accuracy is obtained with batch size 1000, while all other methods operate on batch size 128. Similarly, GGT's best results are achieved with window size  $m = 100$, while for M-FAC larger window sizes improve results; we use $m = 512$. Due to these discrepancies, we measure overheads with the ``best accuracy'' setting for each method but also in the same setup as M-FAC; the latter is given in brackets. 

In terms of accuracy, M-FAC scores highest in the ResNet20 experiment, and comes third, to tuned SGD and AdaHessian (performing almost the same), on ResNet32.
The overhead of M-FAC is around $50\%$, which is lower than AdaHessian, on par with K-FAC, but higher than GGT. However, these overheads are comparable only when GGT uses $5 \times$ less gradients (surprisingly, larger window sizes yield worse results for GGT). On the same settings, M-FAC provides up to 5x lower overhead. This is due to (a) better performance of our dynamic algorithm relative to SVD and (b)  the $\Theta(dm^2)$ term in GGT, versus our $O(dm)$ term. 

\parhead{Image Classification.} Table 4 (right) provides a scenario where we run SGD with momentum, Adam, and M-FAC \emph{without tuning or weight-decay} on Wide Residual Networks (WRN)~\citep{zagoruyko2016wide} for CIFAR-100, as well as on MobileNetV1/ ImageNet, and examine test accuracy. M-FAC achieves the highest accuracy on most runs, even on ImageNet. We emphasize that these results can probably be improved by parameter tuning. Another interesting application  is fine-tuning sparse models. We fine-tune 90\% sparse MobileNet / ResNet50 models for 30 epochs starting from the last gradual pruning step and find that M-FAC reaches higher final accuracy than SGD in both cases. At the same time, it is only negligibly slower ($\leq 5\%$ overhead) since algorithm's complexity (compute and memory) is linear in the model \emph{density}. At last, we note that M-FAC achieving higher test accuracy is generally well correlated with a lower final training loss.

\begin{table}[]
    \centering
    {\small
    \begin{tabular}{clcccccccc}
        \toprule
        & & SQv2 & SST-2 & MRPC & STS-B & QQP & MNLI-m & MNLI-mm & QNLI \\
        \midrule
        t & Adam & 48.41 & 80.11 & 69.90 & 64.39 & 81.09 & 65.36 & 67.78 & 77.85 \\
        t & M-FAC & \textbf{49.80} & \textbf{81.86} & \textbf{72.94} & \textbf{80.15} & \textbf{84.20} & \textbf{68.28} & \textbf{68.98} & \textbf{81.17} \\
        \midrule
        m & Adam & 54.80 & \textbf{85.46} & 76.57 & 82.09 & 86.45 & 73.30 & 74.85 & \textbf{83.85} \\
        m & M-FAC & \textbf{58.02} & 84.20 & \textbf{78.87} & \textbf{84.66} & \textbf{86.75} & \textbf{74.59} & \textbf{75.95} & 83.70 \\
        \bottomrule
    \end{tabular}
    }
    \caption{Average of 5 runs performance of Adam and M-FAC for tiny (t) and mini (m) BERT on SQuADv2 and GLUE benchmark tasks. Due to space, we only show accuracy for SQuADv2 / QQP and the Pearson correlation for STS-B, see Appendix for full results. Models are hosted at \url{https://huggingface.co/M-FAC}.}
    \label{tab:bert-results}
\end{table}

\parhead{Language Modelling using Transformers.} Finally, we test the M-FAC optimizer for smaller Transformer models. We use the default values $m = 1024$ gradients (unless the dataset has less than this number of total samples), dampening $\lambda = 10^{-6}$, learning rate $10^{-4}$ and no weight decay. We then train BERT \cite{devlin2018bert} \textit{tiny} and \textit{mini} models \cite{turc2019well} on SQuADv2 \cite{rajpurkar2018know} and the GLUE benchmark suite \cite{wang2018glue}, comparing against HuggingFace's \cite{wolf-etal-2020-transformers} Adam baseline. For M-FAC, we use exactly the same training setup and only replace the optimizer. The results in Table \ref{tab:bert-results} show that M-FAC performs better than the Adam baseline on almost all tasks, on several even by considerable margins. We provide more detailed results in the Appendix, including hyperparameters and a comparison with AdamW.

\section{Discussion}

We presented static and dynamic algorithms for computing IHVPs when the Hessian matrix can be approximated by a sum of rank-1 matrices. 
We used the classic empirical Fisher approximation, but our results can apply more broadly. 
The main limitation is the cost of storing $m$ additional gradients. 
For the static algorithm, we can efficiently leverage system memory, as described in the Appendix. 
The dynamic algorithm could parallelize gradient storage, or perform gradient compression~\cite{2017-sun}. 
We plan to investigate this in future work and perform a larger-scale study of optimization performance.

\section{Acknowledgements}

We gratefully acknowledge funding the European Research Council (ERC) under the European Union’s Horizon 2020 research and innovation programme (grant agreement No 805223 ScaleML), as well as computational support from Amazon Web Services (AWS) EC2. 

\bibliographystyle{plain}

\newpage

\addtocontents{toc}{\protect\setcounter{tocdepth}{2}}

\newpage
\onecolumn
\appendix

\renewcommand{\contentsname}{Appendix}
\addtocontents{toc}{\protect\setcounter{tocdepth}{2}}
\tableofcontents
\newpage
\begin{center}
	\textbf{\Large{Additional Material}}
\end{center}

\section{Proof of Theorem 1}

We now prove the Theorem 1, which forms the basis for the dynamic algorithm. Before doing that,
we restate the theorem in its full version:

\newtheorem*{theorem-non}{Theorem}
\begin{theorem-non}
    Let $(\nabla \ell_i)_{i = 1}^m$ be a series of gradients, and, for any index $i \geq 1$, let $\widehat{F}_{i} = \lambda I_d + \frac{1}{m} \sum_{j = 1}^i \nabla \ell_j \cdot \nabla \ell_j^\top$ be a dampened version of the empirical Fisher, where $\lambda > 0$ is a small constant. Then $\widehat{F}_{i}^{-1} \vx$ can be calculated as:
    \begin{equation}
        \widehat{F}_{i}^{-1} \vx = \lambda^{-1} \vx - \sum^i_{j=1} c^i_j \nabla \ell_{j}  \,\,\text{ and }\, \,  c^i_j = \sum_{k=j}^{i} \frac{\left(\nabla \ell_k^\top \widehat{F}_{k - 1}^{-1} \vx\right)}{m + [\mathbf{D}]_{kk}} [\mathbf{B}]_{kj},
    \end{equation}
    where $[\mathbf{D}]_{kk} = \nabla \ell_k^\top \widehat{F}_{k - 1}^{-1} \nabla \ell_k$, and $[\mathbf{B}]_{kj}$ being defined such that
    \begin{equation}
        \widehat{F}_{k - 1}^{-1} \nabla \ell_k = \sum_{j = 1}^k [\mathbf{B}]_{kj} \nabla \ell_j \,\, \text{and} \,\, [\mathbf{B}]_{11} = \lambda^{-1}.
    \end{equation}
\end{theorem-non}

\begin{proof}
The proof makes use of the following two equalities:
\begin{align}
    \widehat{F}_{i}^{-1} \vx &= \lambda^{-1} \vx - \sum^i_{j=1} c^i_j \nabla \ell_{j} \label{eq:proof-basis1}  \\
    \widehat{F}_{i}^{-1} \vx &= \lambda^{-1} \vx - \sum_{j = 1}^i \left(\widehat{F}_{j - 1}^{-1} \nabla \ell_j\right) \frac{\left(\widehat{F}_{j - 1}^{-1} \nabla \ell_j\right)^\top \vx}{m + [\mathbf{D}]_{jj}}, \label{eq:proof-basis2} 
\end{align}
which correspond to Equation~\ref{eq:dynamic-idea} with $m = i$ and Equation~\ref{eq:hinv-def} in an unrolled form, respectively. We begin by setting them equal. Next, we apply (\ref{eq:proof-basis1}) again to $\widehat{F}_{j - 1}^{-1} \nabla \ell_j$ (naming the corresponding coefficients $a^{j - 1}_k$) and reorganize the nested sums to get a new expression of the form (\ref{eq:proof-basis1}). Finally, we simplify the result by using our definition of $[\mathbf{B}]_{ij} = -a^{i - 1}_{j}$ for $j < i$ and $[\mathbf{B}]_{ii} = \lambda^{-1}$ (discussed at the beginning of Section~\ref{sec:dynamic}). We obtain: 
\begin{align}
    \widehat{F}_{i}^{-1} \vx 
    &= \lambda^{-1} \vx - \sum^i_{j=1} c^i_j \nabla \ell_{j} \label{eq:proof-1} \\
    &= \lambda^{-1} \vx - \sum_{j = 1}^i \left(\widehat{F}_{j - 1}^{-1} \nabla \ell_j\right) \frac{\left(\widehat{F}_{j - 1}^{-1} \nabla \ell_j\right)^\top \vx}{m + [\mathbf{D}]_{jj}} \\
    &= \lambda^{-1} \vx - \sum_{j = 1}^i \left(\lambda^{-1} \nabla \ell_j - \sum_{k = 1}^{j - 1} a^{j - 1}_k \nabla \ell_k\right) \frac{\left(\nabla \ell_j^\top \widehat{F}_{j - 1}^{-1} \vx\right)}{m + [\mathbf{D}]_{jj}} \\
    &= \lambda^{-1} \vx - \sum_{j = 1}^i \left( \frac{\left(\nabla \ell_j^\top \widehat{F}_{j - 1}^{-1} \vx\right)}{m + [\mathbf{D}]_{jj}} \lambda^{-1} - \sum_{k=j + 1}^{i} \frac{\left(\nabla \ell_k^\top \widehat{F}_{k - 1}^{-1} \vx\right)}{m + [\mathbf{D}]_{kk}} a_j^{k - 1}\right) \nabla \ell_j \\
    &= \lambda^{-1} \vx - \sum_{j = 1}^i \left(\sum_{k=j}^{i} \frac{\left(\nabla \ell_k^\top \widehat{F}_{k - 1}^{-1} \vx\right)}{m + [\mathbf{D}]_{kk}} [\mathbf{B}]_{kj}\right) \nabla \ell_j. \label{eq:proof-2}
\end{align}
We can now directly match coefficients between (\ref{eq:proof-1}) and (\ref{eq:proof-2}), as they hold for any gradient values. This completes the proof.
\end{proof}

\section{Efficiently Implementing the Dynamic Algorithm}

As mentioned in Section~\ref{sec:dynamic}, directly implementing the various recursive formulas of the dynamic algorithm will most likely be quite slow in practice. Thus, we now discuss how to develop an efficient practical implementation. The reader can also find the full code for this implementation of the dynamic algorithm (and the corresponding optimizer) in PyTorch, available at~\cite{Code}.  

\paragraph{Setup.}
We begin by vectorizing the calculations of $\mathbf{D}$ and $\mathbf{B}$. Algorithm~\ref{alg:dynamic-setup} describes the full setup procedure in PyTorch-like pseudocode.

\begin{algorithm}[h]
    \caption{Calculating $\mathbf{D}$ and $\mathbf{B}$ in $O(m^3)$ time and $O(m^2)$ memory, assuming that $\mathbf{G}\mathbf{G^\top}$ is already precomputed.}
    \label{alg:dynamic-setup}
    \begin{algorithmic}
         \STATE $\mathbf{D} \gets \lambda^{-1} \mathbf{G}\mathbf{G^\top}$
         \STATE $\mathbf{B} \gets \lambda^{-1} \mathbf{I}_{m \times m}$
         \vspace{10pt}
         \FOR {$i \gets 2, 3, \dots, m$}
             \STATE $\mathbf{D}_{i:, i:} \gets \mathbf{D}_{i:, i:} -  \frac{1}{m + D_{i - 1, i - 1}}  \mathbf{D}_{i-1:, i:}^\top \mathbf{D}_{i-1:, i:}$
         \ENDFOR
         \STATE $\mathbf{D} \gets \text{triu}(\mathbf{D})$
         \vspace{10pt}
         \FOR {$i \gets 2, 3, \dots, m$}
             \STATE $\mathbf{B}_{i, :i} \gets  (-\mathbf{D}_{:i, i} / (m + \text{diag}(\mathbf{D})_{:i}))^\top \mathbf{B}_{:i, :i}$
         \ENDFOR
     \end{algorithmic}
\end{algorithm}

\begin{figure}[h]
    \centering
    \includegraphics[width=\textwidth]{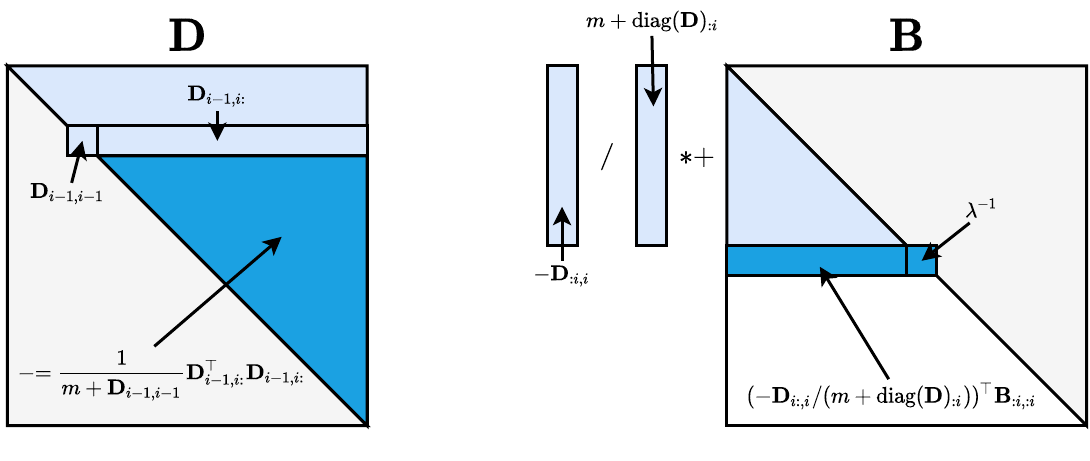}
    \caption{Visualized calculation of $\mathbf{D}$ and $\mathbf{B}$.}
    \label{fig:dynamic-vis}
\end{figure}

Initially, $\mathbf{D}$ is the stored scalar product matrix $\mathbf{G}\mathbf{G}^\top$ scaled by $\lambda^{-1}$. Row 1 already has the correct final values. The algorithm then goes through $m - 1$ iterations, each completing one row but also updating all the rows of higher index. More concretely, in iteration $i$, we subtract the outer product of the previous row starting at the element one after the diagonal $\mathbf{D}_{{i - 1}, i:}$ with itself, scaled by the inverse of $m$ plus the previous diagonal element $\mathbf{D}_{i - 1, i - 1}$, from the square sub-matrix with $ii$ as the upper left corner, i.e. $\mathbf{D}_{i:, i:}$. This process is also visualized in Figure~\ref{fig:dynamic-vis}, where the already calculated elements are shaded in light blue, the ones that are currently being updated in darker blue, and the irrelevant ones in grey. Since the calculation of $\mathbf{D}$, as given here, will also produce some unnecessary (and incorrect) values below the diagonal, we delete those after the loop by extracting just the upper triangular part of $\mathbf{D}$. $\mathbf{B}$ starts off as the identity matrix times $\lambda^{-1}$ (and row 1 is again already done). Next, $\mathbf{B}$ is calculated row by row, where row $i$ is a negative linear combination of the previous $i - 1$ rows (up to index $i - 1$ as $\mathbf{B}$ is lower triangular) $\mathbf{B}_{:i, :i}$ with the coefficients given by the first $i - 1$ elements of column $i$ of $\mathbf{D}$, i.e. $\mathbf{D}_{:i, i}$, each divided by $m$ plus the corresponding diagonal element of $\mathbf{D}$, i.e. $m + \text{diag}(\mathbf{D})_i$. This process is also shown in Figure~\ref{fig:dynamic-vis}.

In theory, the algorithm presented so far should be well suited for utilizing the massive parallel computation capabilities of a GPU. However, a straightforward implementation in e.g. PyTorch, will not be very fast for larger values of $m$ as it results in thousands of small kernel launches with overhead that adds up to a very significant overall slow-down. (Concretely, we measured $> 7 \times$ overhead for $m = 1024$). This means that, to achieve good performance, the setup for the coefficient computation of the dynamic algorithm needs to be implemented with (custom) CUDA kernels that merge computations to get rid of the excessive kernel launch overhead.

Carefully studying the calculation process of $\mathbf{D}$ reveals that it is actually very similar to a Cholesky decomposition (concretely, the only difference is that it does not use the square root of the diagonal element but the element itself plus $m$). In fact, it is actually exactly equivalent to taking the upper triangular matrix in the LU-decomposition (with no pivoting) of $\mathbf{\lambda^{-1} GG^\top} + m \mathbf{I}$ and then subtracting $m \mathbf{I}$ again. This means that it is possible to reuse the highly optimized LU-decomposition kernel provided by PyTorch to calculate $\mathbf{D}$ quite efficiently (a custom kernel implementing the modified Cholesky decomposition would certainly be faster, but we already observed pretty low overheads with the LU version). For the efficient computation of $\mathbf{B}$, a custom kernel is needed, which we provide at~\cite{Code}, and describe next. 

\begin{figure}[h]
    \centering
    \includegraphics[width=.3\textwidth]{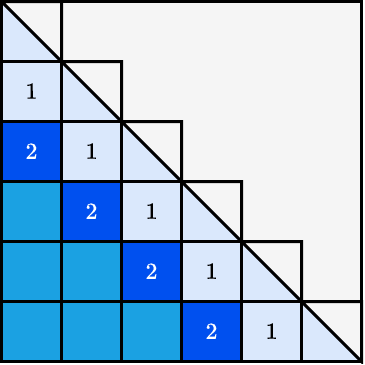}
    \caption{Visualization of the CUDA kernel for calculating $\mathbf{B}$. In each iteration, all blocks below the previous diagonal (here denoted as 1) are updated in parallel and those adjacent to the diagonal (here denoted as 2) are completed to form the reference diagonal for the next iteration.}
    \label{fig:dynamic-cuda}
\end{figure}

The main idea is that we can dramatically reduce the number of separate kernel launches by computing multiple rows of $\mathbf{B}$ in a single kernel. In general, we split $\mathbf{B}$ into blocks of $32 \times 32$ (this matches the fact that modern NVIDIA GPUs have 1024 threads per block). Then, we begin by fully computing all diagonal blocks in parallel. Notice that, due to the lower triangular structure of $\mathbf{B}$, those are fully independent. Similarly, each block depends exclusively on the blocks above it (up to the diagonal). Next, we perform an iterative process which will update all remaining blocks with respect to the new values of the most recently computed diagonal before completing the calculation of the blocks adjacent to it. This new diagonal is then the reference for the next iteration. Figure~\ref{fig:dynamic-cuda} visualizes this process. The calculation within the individual blocks is easy to parallelize. For more details, please refer to our code for this CUDA kernel. Overall, this  way of calculating $\mathbf{B}$ is quite fast; e.g. for $m = 1024$, it takes $< 2$ milliseconds to execute on an NVIDIA RTX 3090 GPU.

\paragraph{IHVPs}

After discussing an efficient implementation of the precomputation steps, we now focus on actually computing IHVPs. Algorithm~\ref{alg:dynamic-mul} presents a vectorized implementation.

\begin{algorithm}[h]
    \caption{Computing $\widehat{F}_{m}^{-1} \vx$ in $O(dm + m^2)$ time, assuming that $\mathbf{D}$ and $\mathbf{B}$ have already been precomputed.}
     \label{alg:dynamic-mul}
     \begin{algorithmic}
         \STATE $\mathbf{q} \gets \lambda^{-1} \mathbf{G}\vx$
         \STATE $q_1 \gets q_1 / (m + \mathbf{D}_{11})$
         \FOR {$i \gets 2, 3, \dots, m$}
             \STATE $\vq_{i:} \gets \vq_{i:} - q_{i - 1} \mathbf{D}_{i - 1, i:}^\top$
         \ENDFOR
         \STATE $\mathbf{q} \gets \mathbf{q} / (m + \text{diag}(\mathbf{D}))$
         \STATE \textbf{return} $\lambda^{-1} \vx - ((\vq^\top \mathbf{B}) \mathbf{G})^\top$
    \end{algorithmic}
\end{algorithm}

We note that if $\mathbf{x} = \nabla \ell_i$, then $\mathbf{q}_{:(i + 1)} = \mathbf{D}_{:(i + 1), i} / \text{diag}(\mathbf{D}_{:(i + 1)})$, i.e. the first $i$ elements in column $i$ of $\mathbf{D}$ divided element-wise by the corresponding diagonal values. This can be utilized for a more efficient combined update-multiply operation, which is essential for the M-FAC optimizer, where we first update our Hessian estimation with a gradient $\nabla \ell$ and then return the IVHP with this same $\nabla \ell$. Further, reusing the same $\mathbf{p} = \mathbf{G} \nabla \ell$ for both updating $\mathbf{G}\mathbf{G}^\top$ as well as calculating $\mathbf{q}$, saves one redundant $O(dm)$ matrix-vector-product.

\section{Efficiently Implementing the Static Algorithm}

Although the formulas discussed in the main body of the paper have low theoretical time complexity, a direct implementation does not make the best use of modern machine learning frameworks and GPUs. Therefore, we now present more efficient matrix-based versions. We note that a block-wise version would simply apply the techniques discussed here independently to each block.

Let $\mathbf{V} = [\mathbf{v_1}^\top; \mathbf{v_2}^\top; \dots; \mathbf{v_m}^\top]$ be the $m \times d$ row-wise matrix of the $\mathbf{v_i}$ and $\mathbf{q} = (q_1, q_2, \dots, q_m)^\top$ the $m$-vector of the $q_i$. Then, $\widehat{F}_{m}^{-1} \vx$ can be written as follows, where $/$ denotes the element-wise division:
\begin{equation}
    \label{eq:v1-efficient-mul}
    \widehat{F}_{m}^{-1} \vx = \lambda^{-1} \vx - \mathbf{V}^\top \left(\left(\mathbf{V}\vx\right) / \vq\right).
\end{equation}
It is also possible to extract one element of each matrix row simultaneously (in terms of vectorized operations) with $O(dm)$ cost, for example the entire diagonal. Let $\boldsymbol{\pi}(\vx)$ be a mapping such that $\boldsymbol{\pi}(\vx)_i = x_j$ where $j$ is the column of the element to select in row $i$, e.g. $\boldsymbol{\pi}(\vx) = \vx$ to get the diagonal, and $\mathbf{e^\pi}$ a vector such that $e^\pi_i = 1$ if $\pi(\vx)_i = x_i$ and $e^\pi_i = 0$ otherwise. Using these definitions, the calculation of the desired result $\vy$ is described below, where $\odot$ denotes the element-wise product.
\begin{equation}
    \vy = \lambda^{-1} \mathbf{e^\pi} - \sum^m_{i = 1} \frac{1}{q_k} \mathbf{v_i} \odot \boldsymbol{\pi}(\mathbf{v_i})
\end{equation}
Our implementation contains several additional optimizations. 
For instance, we can efficiently precompute $\mathbf{V}$ and $\mathbf{q}$ by repeatedly applying (\ref{eq:v1-efficient-mul}) to produce the next $\mathbf{v_i}$. In the next section, we additionally discuss several memory-saving optimizations, and in particular explicit page swapping between CPU and GPU memory for situations where $\mathbf{V}$ does not fully fit in the GPU memory. 

Finally, Algorithm~\ref{alg:static-setup} demonstrates, using a Python-like matrix indexing syntax, how to efficiently precompute $\mathbf{V}$ and $\mathbf{q}$ by repeatedly applying (\ref{eq:v1-efficient-mul}) to produce the next $\mathbf{v_i}$. It should be noted that the row-wise matrix of gradients $\mathbf{G} = [\nabla \ell_1^\top; \nabla \ell_2^\top; \dots; \nabla \ell_m^\top]$ is re-purposed as $\mathbf{V}$, a simple way to halve the peak memory consumption in practice.

\begin{algorithm}
    \caption{Precomputation of $\mathbf{V}$ and $\vq$ given $\mathbf{G}$.}
    \label{alg:static-setup}
    \begin{algorithmic}
        \STATE $\mathbf{V} \gets \mathbf{G}$
        \STATE $\vg \gets \mathbf{V}_{1, :}^\top$
        \STATE $\mathbf{V}_{1, :} \gets \lambda^{-1} \vg$
        \STATE $q_1 \gets \mathbf{V}_{1, :} \, \vg$
        \FOR {$i \gets 2, 3 \dots, m$}
            \STATE $\vg \gets \mathbf{V}_{i, :}^\top$
            \STATE $\mathbf{V}_{i, :} \gets \lambda^{-1} \vg^\top - ((\mathbf{V}_{:i, :} \, \vg) / \vq)^\top \mathbf{V}_{:i, :}$
            \STATE $q_i \gets m + \mathbf{V}_{i, :} \, \vg$
        \ENDFOR
    \end{algorithmic}
\end{algorithm}

\subsection{Additional Optimizations}

\begin{figure}[h]
    \centering
    \includegraphics[width=.75\textwidth]{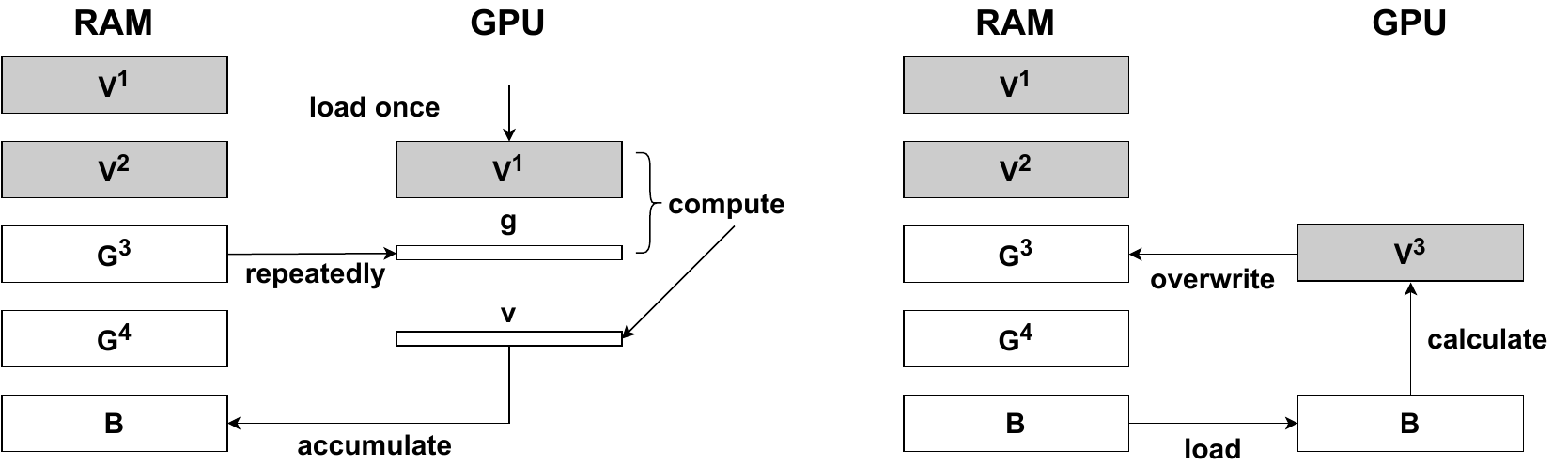}
    \caption{Visualization of the swapping implementation of the static algorithm, grey blocks have already been fully calculated. \textbf{Left}: $g$ is loaded from block $\mathbf{G}^3$, which allows calculating $\vv$ using the GPU block $\mathbf{V}^1$ and then accumulating it into the buffer $\mathbf{B}$. \textbf{Right}: Calculation of $\mathbf{V}^3$ is finished using the completed buffer.}
    \label{fig:swapping}
\end{figure}

As we have seen, obtaining a good Fisher approximation of the Hessian requires a sizable number of gradients, which means that, for bigger networks, it can easily happen that the collection of all gradients used for Fisher estimation does not fully fit into GPU memory. We now discuss how to efficiently handle such situations, with an implementation that performs explicit swapping of gradient blocks between GPU and CPU memory (RAM). The most important steps of the method to be discussed are also visualized in Figure~\ref{fig:swapping}.

In general, a simple trick to halve the peak memory consumption of the static algorithm is repurposing the gradient matrix $\mathbf{G}$ as $\mathbf{V}$, which is possible as $\nabla \ell_i$ is not needed anymore after $\vv_i$ and $q_i$ have been calculated. Now, for the explicit swapping implementation, we first split the collection of $m$ gradients into $k$ blocks / pages $\mathbf{G}^i$ containing at most $m / k$ gradients each. Those are then turned into the corresponding $\mathbf{V}^i$ blocks in increasing order of $i$. Additionally, we maintain a single buffer block $\mathbf{B}$ of the same size for accumulating intermediate results. All blocks reside in CPU memory and are only loaded to the GPU as needed, meanwhile $\vq$ is small enough to be kept in GPU memory at all times. To compute $\mathbf{V}^i$ we first load block $\mathbf{V}^1$ fully into GPU memory. Then we load the first gradient of $\mathbf{G}^i$ denoted by $\mathbf{g}^i_1$ and compute the corresponding $\vv^1_1$ with respect to the loaded $\mathbf{V}^1$, which is afterwards saved in the first index of the buffer $\vb_1$. After repeating this for all $\vg^i_j$, block $\mathbf{V}^1$ is swapped with block $\mathbf{V}^2$ and the whole process starts again, accumulating the resulting $\vv^2_j$ into the buffer. Eventually, after handling $\mathbf{V}^{i - 1}$, we can load $\mathbf{B}$ into GPU memory, finish the calculation of the $\vv^i_j$ and store them in $\mathbf{V}^i$ (reusing the memory of $\mathbf{G}^i$). It should be noted that the loading of $\vg^i_j$ can be parallelized with the calculation of $\vg^i_{j-1}$ and thus costs almost no extra time. Finally, one wants to choose the number of blocks $k$ to be as small as possible, to minimize the overhead caused by the $O(k^2)$ page swaps.

Overall, the  implementation described above is effective in practice and allows scaling up our static algorithm (with only modest overhead over a pure GPU implementation), in its full non-blockwise version, to a large number of gradients, even for relatively large networks such as ResNet50.

Lastly, we note that, for a block-wise implementation, where a single block and all its corresponding gradient parts, easily fit into GPU memory (i.e. the opposite of the situation described thus far in this section), it can be beneficial to handle multiple blocks simultaneously via batch-matrix-multiplies. We provide an implementation that can do this, leading to very significant speed-ups on lower block-sizes where otherwise most of the time would be spent constantly loading in new memory from the CPU.

\section{Additional Experimental Results}

\subsection{Pruning Update Approximation}
\label{app:update-approximation}

As discussed in the main paper, we use the OBS pruning framework but prune multiple weights at the same time, which as pointed out by \cite{2019-wang}, in extreme cases, is not guaranteed to be better than OBD due to not taking into account potential correlations. Thus, we now explore how much of a problem this is in practice. To do that, we compare three different M-FAC variations: not updating the non-pruned weights at all (i.e. OBD, but with a better Hessian approximation than the diagonal), jointly applying the OBS update for all pruned weights (i.e. the OBS approximation) and applying the correct simultaneous update by solving a large linear system (see \cite{2020-singh} for details). We prune a ResNet20 / CIFAR-10 model in steps of 10\% down to 90\% sparsity, using three variations of the full block size M-FAC with $m = 1024$ while recomputing the inverse Hessian approximation after every pruning step. The recomputations ensure that we have a reasonably accurate loss approximation at every pruning step and that the number of weights pruned at each step is small enough to solve the corresponding linear system. Table~\ref{tab:update} provides the results. 

\begin{table}[h]
    \centering
    \begin{tabular}{lccccccccc}
         Method & 10\% & 20\% & 30\% & 40\% & 50\% & 60\% & 70\% & 80\% & 90\% \\
         \toprule
         No update (OBD) & 91.4 & 91.3 & 91.0 & 90.2 & 88.4 & 84.8 & 77.2 & 50.4 & 11.5 \\
         Simultaneous OBS & 91.4 & 91.4 & 91.5 & 91.2 & 90.8 & 89.7 & 86.8 & 75.6 & 28.2 \\
         Linear solving & 91.5 & 91.4 & 91.4 & 91.2 & 90.7 & 89.7 & 87.5 & 75.7 & 28.1 \\
         \bottomrule
    \end{tabular}
    \caption{Comparing ResNet20 / CIFAR-10 accuracies of different M-FAC variations at varying levels of sparsity.}
    \label{tab:update}
\end{table}

The results show a very clear gap in accuracy between the no-update and the approximate update version. At the same time, there appears to be almost no difference between the simultaneous OBS and the true OBS update, which has to solve a (potentially very large) linear system. This suggests that the simultaneous update approximation done for computational tractability is also a reasonable choice in terms of achieved accuracy.

\subsection{Ablation Studies for One-Shot Pruning}

To further examine the properties of the local loss approximation, we now present ablation studies on pretrained ResNet50/ImageNet and MobileNetV1-STR/ImageNet models. 
Experiments perform one-shot pruning according to the OBS metric estimated using M-FAC, with varying block size and number of gradients. The dampening factor $\lambda$ in these experiments is set to $10^{-5}$. Following~\citep{2020-singh}, we used batched gradients (of size 16) as single samples inside the Fisher approximation. (This does not alter results, but reduces variance.) 

The goal of these experiments is to examine two questions: 1) does larger block size always imply a better approximation and 2) does higher number of gradients always imply a better approximation? 
We will see that neither of these questions have obvious answers.

\textit{The numbers presented for our one-shot experiments are the averages over two runs. As the variance is very small, we omit error bars.} We sometimes break the standard convention and ``zoom into'' the y axis for visibility.

\subsubsection{ResNet50 / ImageNet}

\begin{figure}[!h]
    \centering
    \begin{subfigure}[t]{0.45\textwidth}
        \centering
        \includegraphics[width=\textwidth]{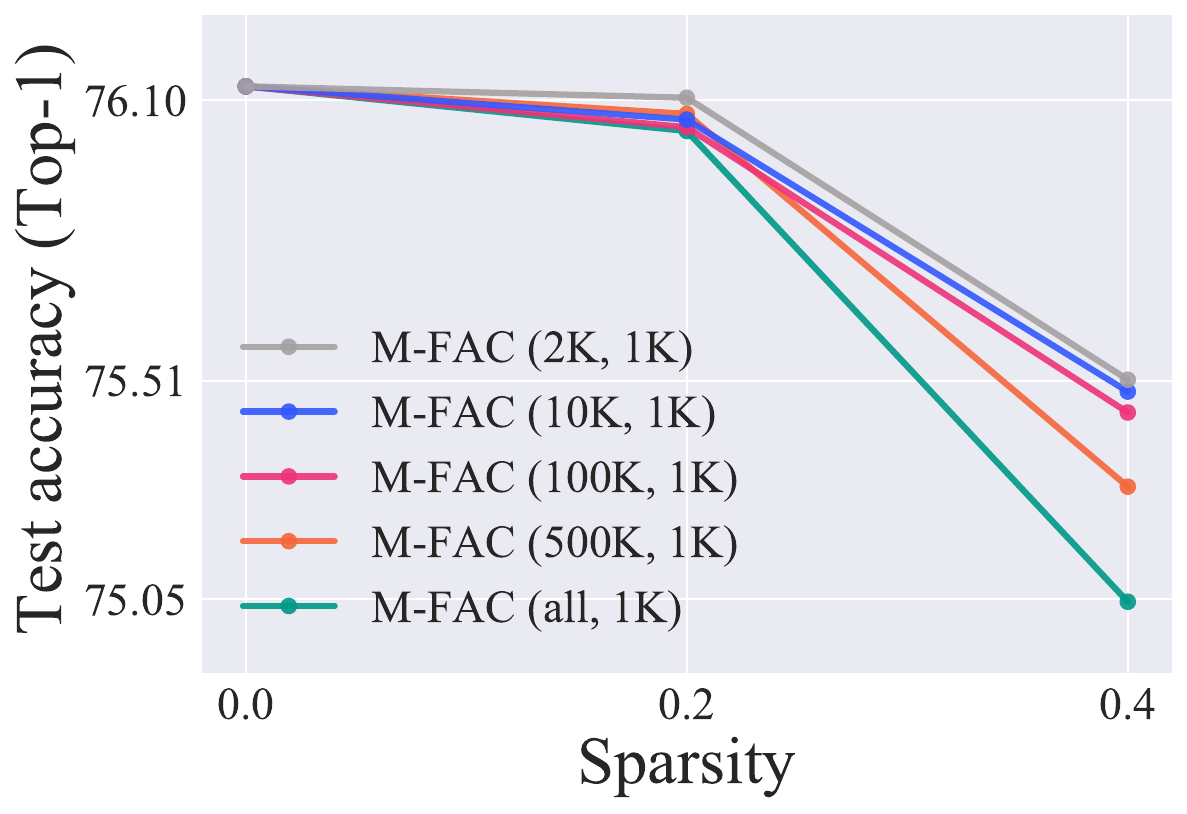}
    \end{subfigure}
    \hfill
    \begin{subfigure}[t]{0.45\textwidth}
        \centering
        \includegraphics[width=\textwidth]{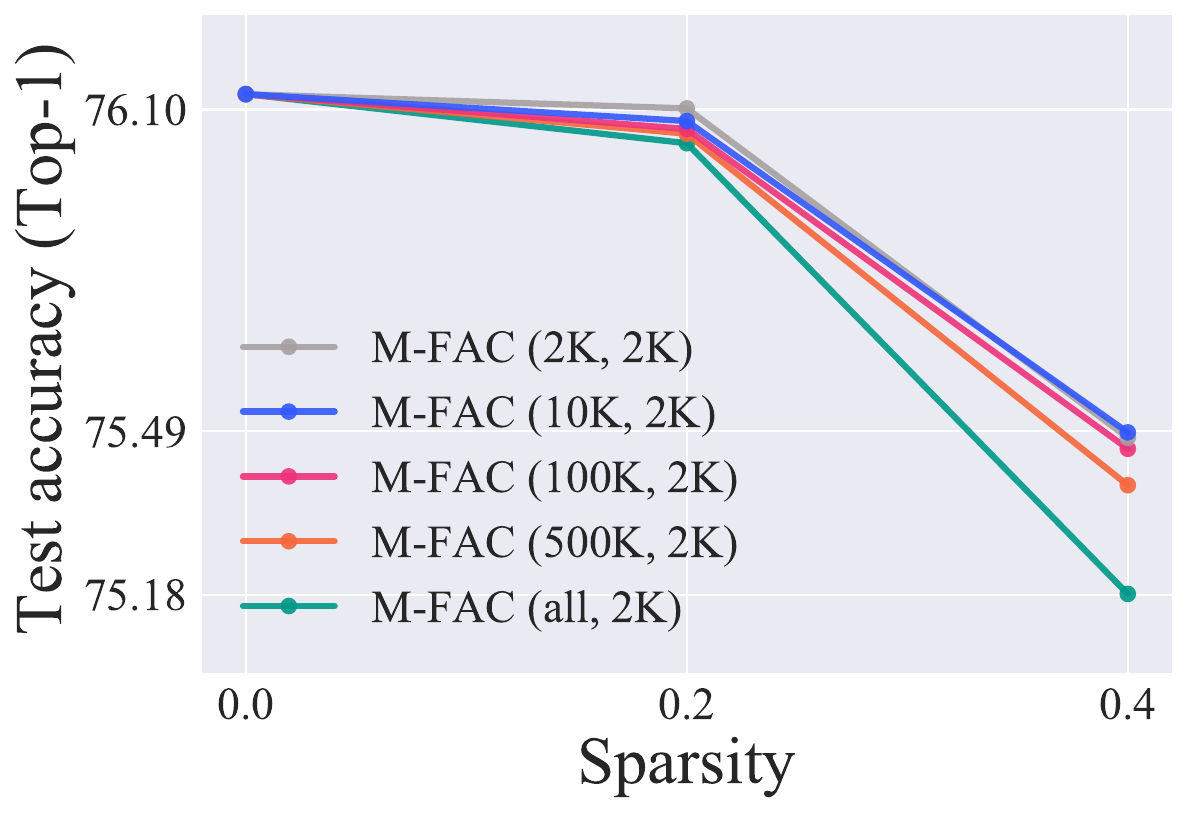}
    \end{subfigure}
    \caption{An ablation study on the ResNet50 / ImageNet model showing the effect of the block size for a fixed number of gradients.}
    \label{fig:rn50-ablation-grad}
    \vspace{-0.5em}
\end{figure}

The first set of experiments examines the dependency on the number of gradients and block size for the ResNet50 / ImageNet model. 
The left subfigure in Figure~\ref{fig:rn50-ablation-grad} shows results for block sizes between $2K$ and all weights, i.e. $d$, for a fixed number of $1K$ gradients, while the right subfigure shows the same, but for $2K$ gradients. 
The first graph presents a fairly unintuitive finding, i.e. that \emph{lower} block sizes appear to provide better pruning accuracy. 
We analyze this finding in detail and explain it in Section~\ref{sec:oneshot-discussion}: roughly, this is due to the fact that gradient entries are scaled by the gradient norm over the block.
As predicted by our analysis, this effect is mitigated by increasing the number of gradients used for estimation, in which case block size $10K$ yields the best results, and the performance of full block size also improves. Please see Section~\ref{sec:oneshot-discussion} for the full analysis. 

\begin{figure}[!h]
    \centering
    \begin{subfigure}[t]{0.45\textwidth}
        \centering
        \includegraphics[width=\textwidth]{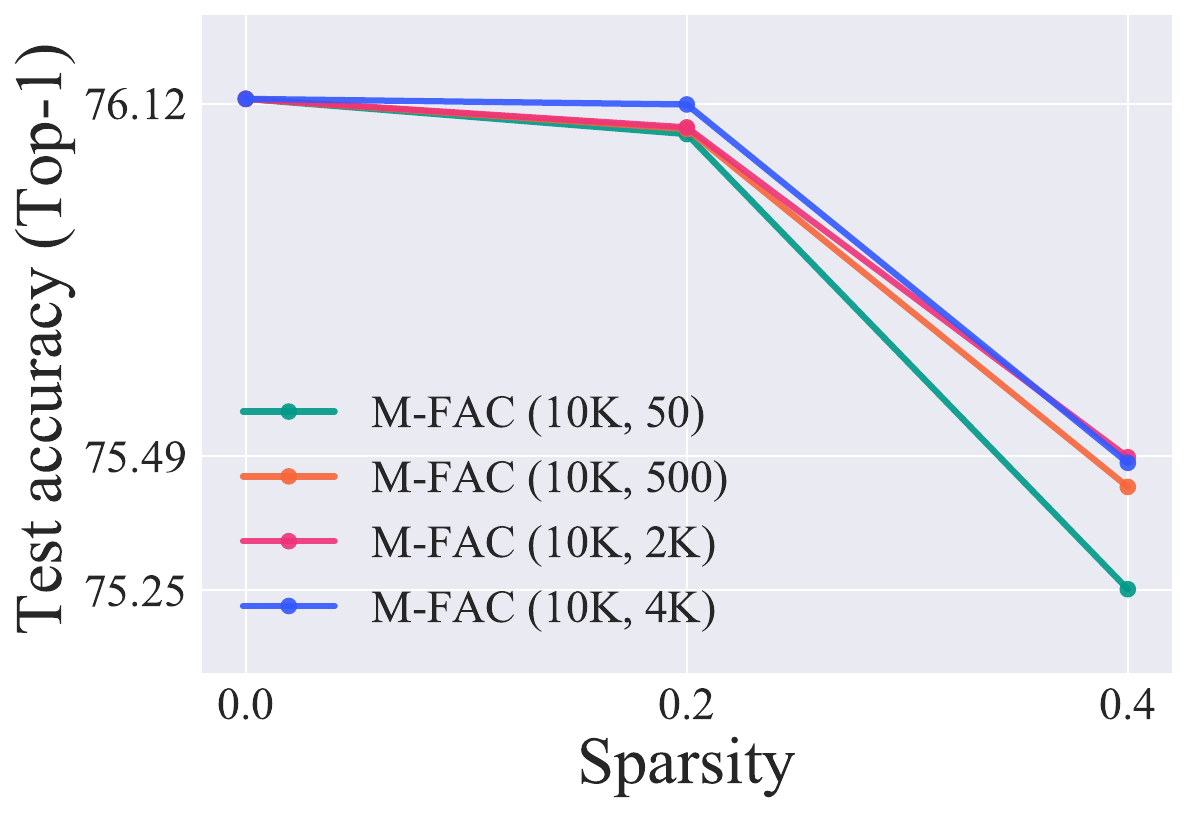}
    \end{subfigure}
    \hfill
    \begin{subfigure}[t]{0.45\textwidth}
        \centering
        \includegraphics[width=\textwidth]{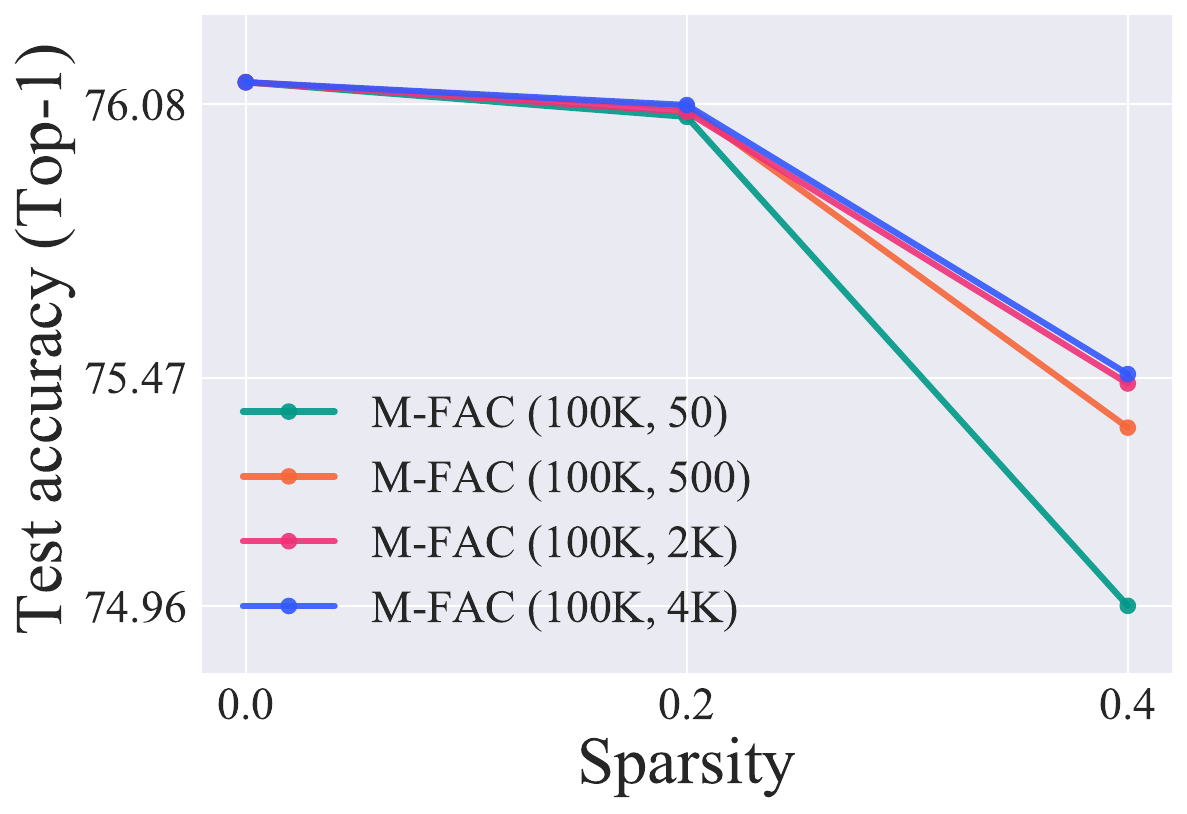}
    \end{subfigure}
    \caption{An ablation study on the ResNet50 / ImageNet model showing the effect of the number of gradients for a fixed block size.}
    \label{fig:rn50-ablation-block}
    \vspace{-0.5em}
\end{figure}

Figure~\ref{fig:rn50-ablation-block} examines the complementary effect, that of number of gradients for a fixed block size (10K and 100K, respectively). The results suggest that more gradients help improve the estimation of the Fisher matrix, although we observe a saturation effect. We also note that, at higher one-shot sparsities (e.g., 60\% in one-shot, not shown for visibility), this effect does not always hold. However, for such large pruning ``perturbations'' the pruning assumption that the Hessian is constant along the direction of pruning is unlikely to hold, which affects the stability of the results.

\subsubsection{MobileNetV1-STR / ImageNet}

\begin{figure}[!h]
    \centering
    \begin{subfigure}[t]{0.45\textwidth}
        \centering
        \includegraphics[width=\textwidth]{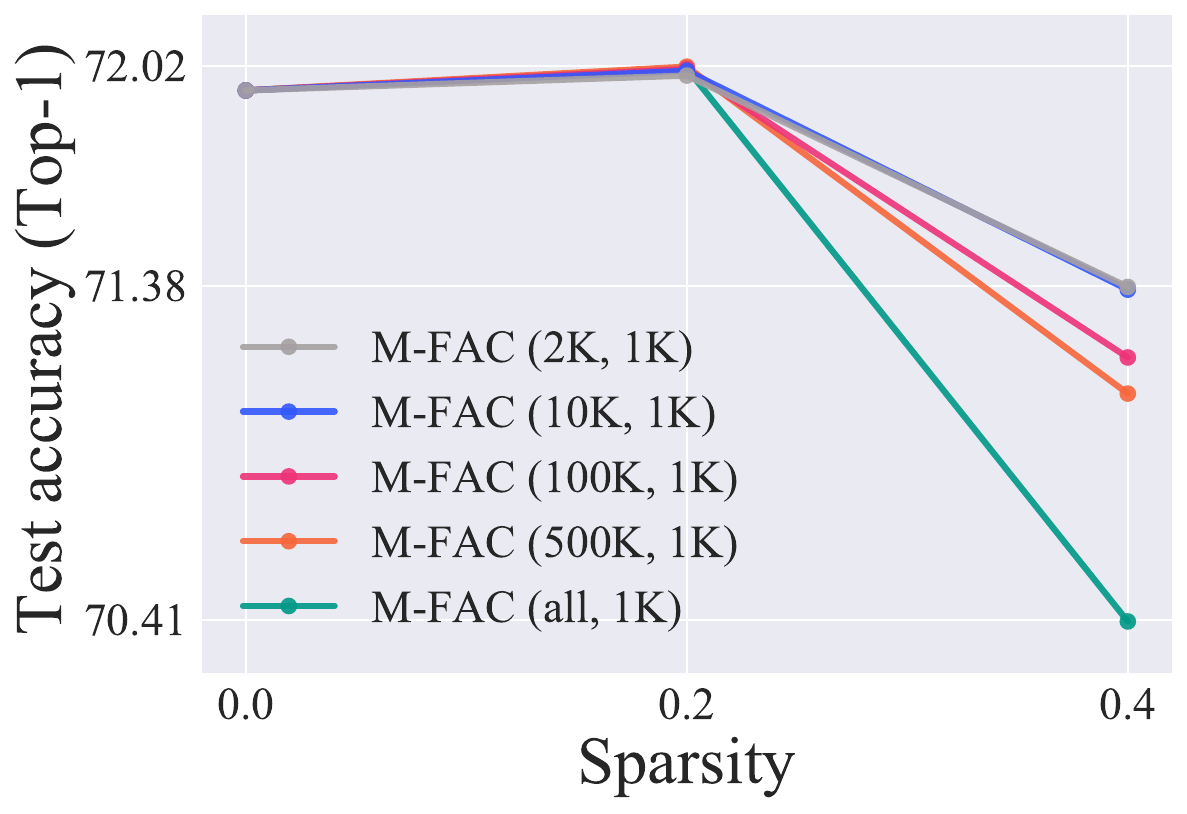}
    \end{subfigure}
    \hfill
    \begin{subfigure}[t]{0.45\textwidth}
        \centering
        \includegraphics[width=\textwidth]{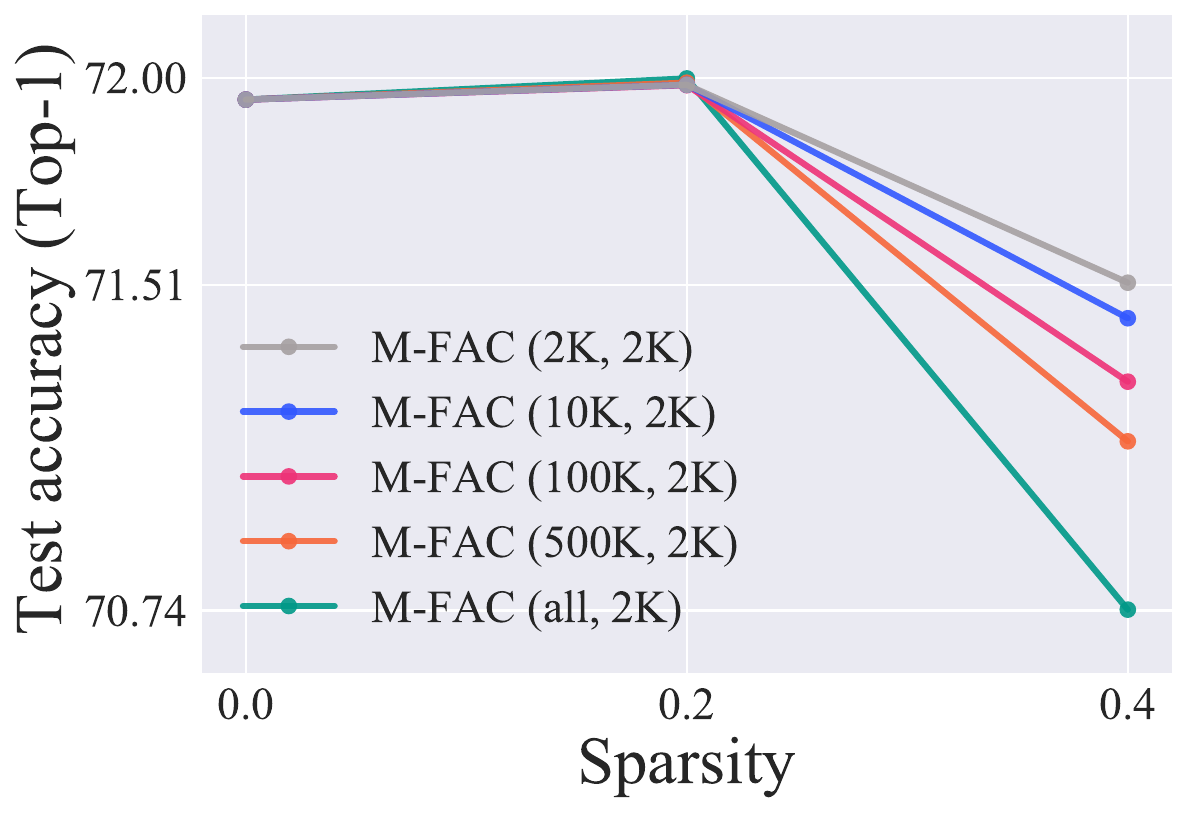}
    \end{subfigure}
    \caption{An ablation study on the MobileNetV1-STR / ImageNet model showing the effect of the block size for a fixed number of gradients.}
    \label{fig:mbv1-ablation-grad}
    \vspace{-0.5em}
\end{figure}

The second set of experiments shows the dependency on the number of gradients and block size for the MobileNetV1-STR / ImageNet model. We use the implementation and pre-trained weights from~\citep{2020-kusupati}. 
Figure~\ref{fig:mbv1-ablation-grad} shows results for block sizes between $2K$ and all weights, i.e. $d$, for a fixed number of gradients ($1K$ in the left subfigure and $2K$ in the right subfigure). 
Again, in both cases, for this model \emph{lower} block sizes appear to provide an improved approximation. 
This is most likely due to the gradient scaling effects which we discuss in Section~\ref{sec:oneshot-discussion}. However, the results show that for this compact model these effects are more prevalent than for the ResNet50 model.

\begin{figure}[!h]
    \centering
    \begin{subfigure}[t]{0.45\textwidth}
        \centering
        \includegraphics[width=\textwidth]{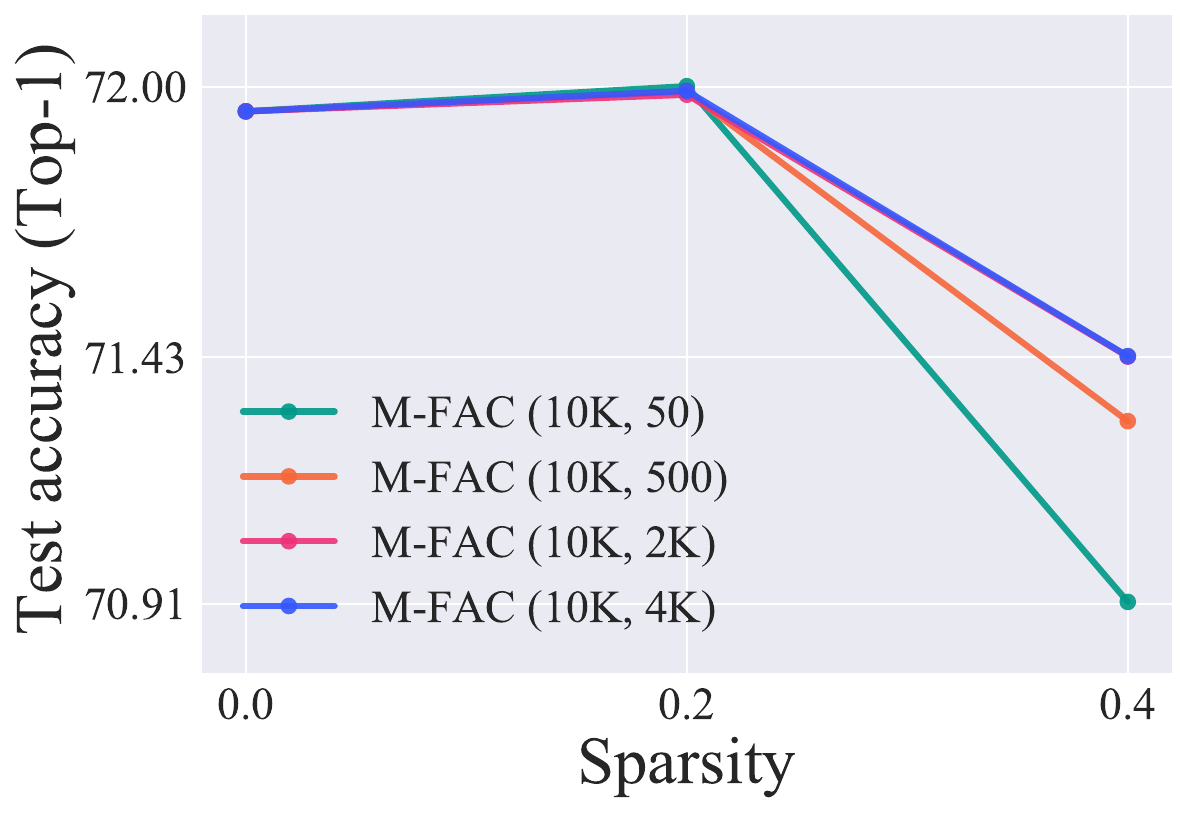}
    \end{subfigure}
    \hfill
    \begin{subfigure}[t]{0.45\textwidth}
        \centering
        \includegraphics[width=\textwidth]{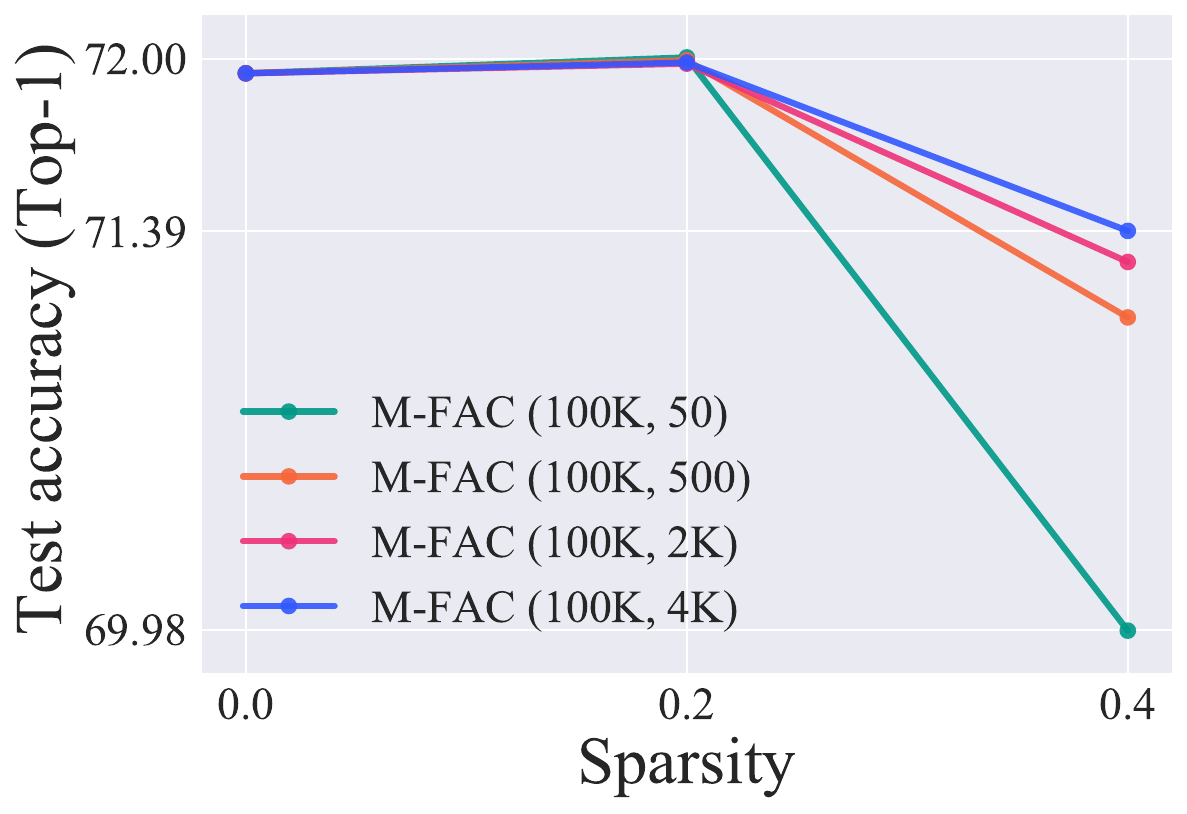}
    \end{subfigure}
    \caption{An ablation study on the MobileNetV1-STR / ImageNet model showing the effect of the number of gradients for a fixed block size.}
    \label{fig:mbv1-ablation-block}
    \vspace{-0.5em}
\end{figure}

Figure~\ref{fig:mbv1-ablation-block} examines the opposite effect, that of the number of gradients for a fixed block size ($10K$ in the left subfigure and $100K$ in the right subfigure). We show results for the number of gradients varying between $50$ and $4K$.
The results clearly suggest that more gradients help improve the accuracy, although the improvement appears to saturate, e.g. between $2K$ and $4K$ gradients.

\vspace{-1em}
\vspace{-1em}

\vspace{20pt}

\subsubsection{Normalizer-Free Nets}

We now examine the ``compressibility'' of the recently-proposed normalizer-free nets, which have shown competitive accuracy based on a similar structure to standard residual networks but without the batch normalization component~\citep{brock2021high}. We use the PyTorch re-implementation of~\citep{rw2019timm}. 
In Figure~\ref{fig:rn50-vs-nfrn50}, we provide a relative comparison in terms of pruned accuracy between a normalizer-free and a regular version of ResNet50. These two networks have virtually the same number of parameters (approximately 25M); however, we notice that the normalizer-free variant is significantly ``easier'' to prune, in terms of the relative accuracy drop versus the dense baseline. We conjecture that this is because of the elimination of the \textit{BatchNorm} layers. 
Specifically, when performing large one-shot pruning steps, the BatchNorm statistics become invalid following a pruning step, which can lead to a significant loss of accuracy; moreover, their removal may render the Fisher approximation more accurate, as the model loss is more stable in the local neighborhood.

\subsubsection{YOLOv3}

\begin{figure}[h]
    \centering
        \begin{subfigure}[t]{0.45\textwidth}
        \centering
        \includegraphics[width=\linewidth]{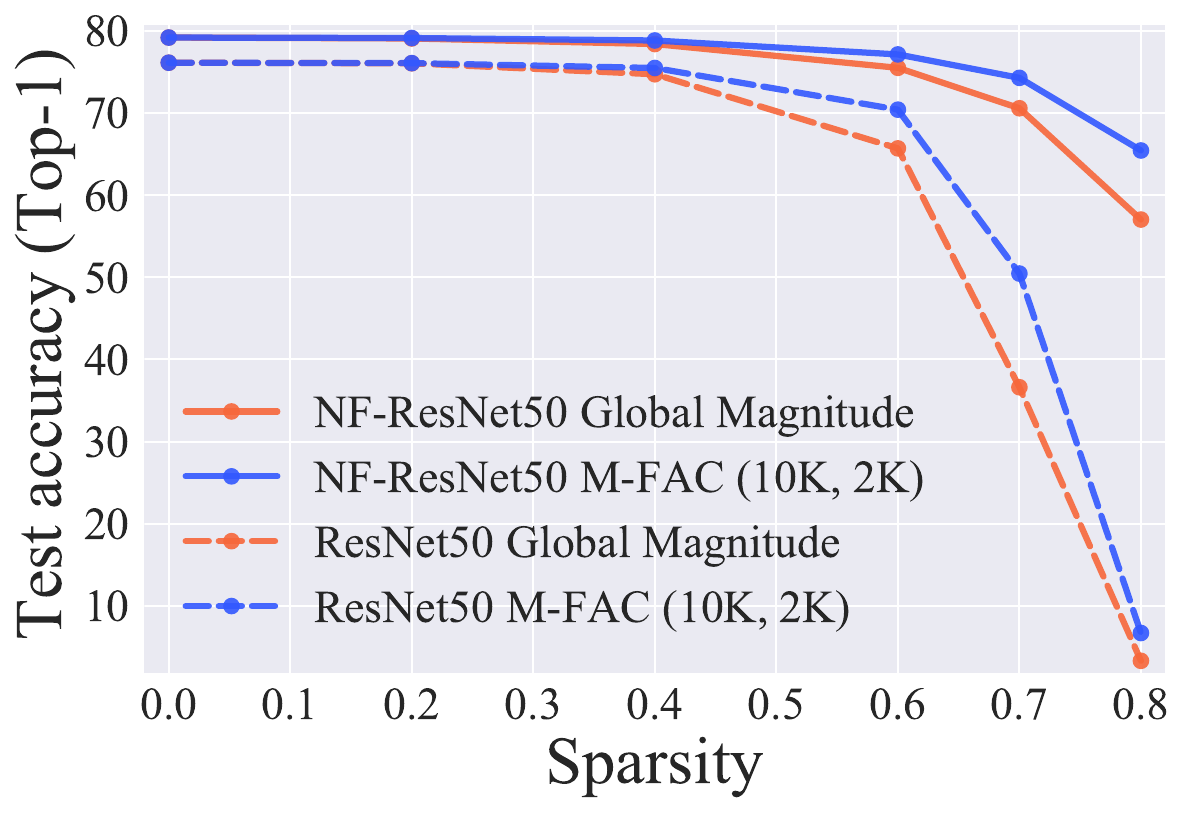}
        \caption{Comparison between GMP and M-FAC pruning on the standard ResNet50 and a Normalizer-Free variant (NF-ResNet50). Observe the significantly better accuracy given by our method on the normalizer-free variant.}
        \label{fig:rn50-vs-nfrn50}
    \end{subfigure}
    \hfill
        \begin{subfigure}[t]{0.45\textwidth}
        \centering
        \includegraphics[width=\linewidth]{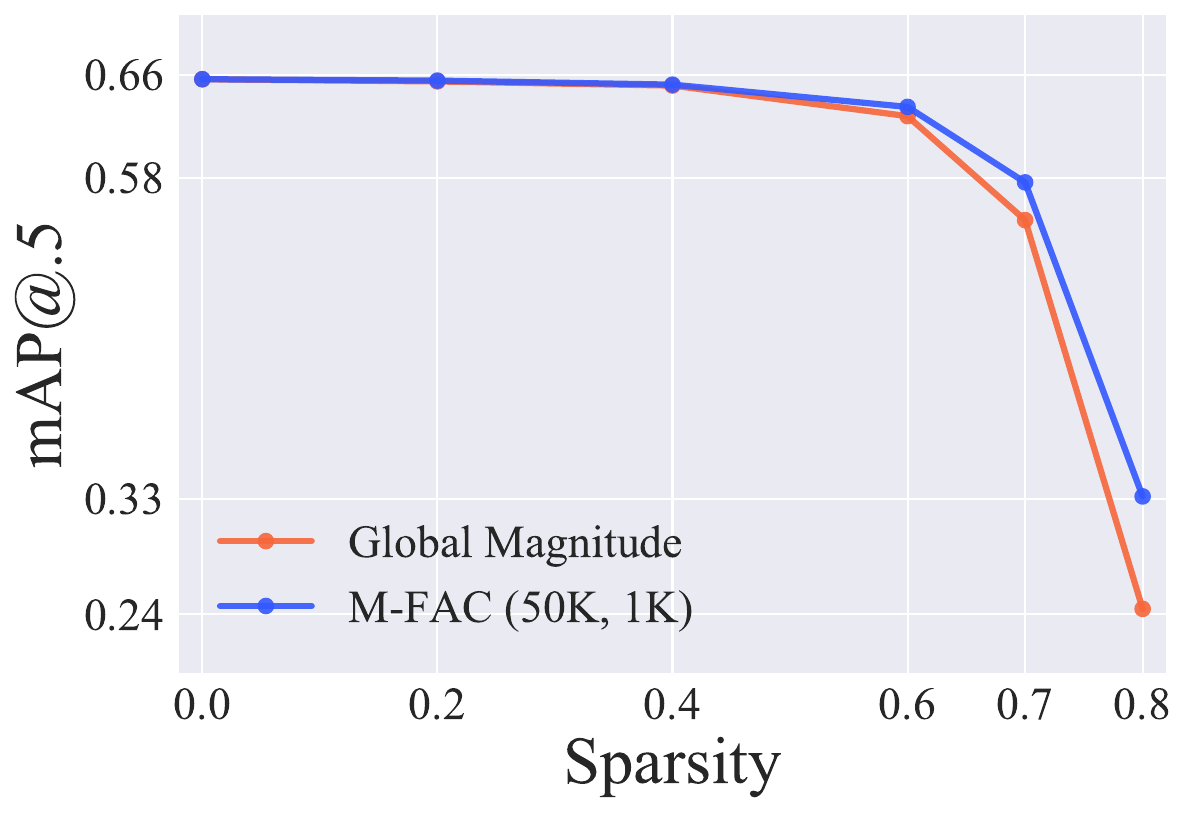}
        \caption{Comparison between GMP and M-FAC in terms of the mAP@0.5 metric for a one-shot pruned complex  YOLOv3-SPP / COCO model.}
        \label{fig:yolov3}
    \end{subfigure}
    \caption{\textbf{Left}: The effectiveness of M-FAC pruning on Normalizer-Free ResNet50 versus the regular variant. \textbf{Right}: One-shot pruning comparison for the YOLOv3-SPP model on the large-scale object detection dataset COCO.}
\end{figure}

Next, we study the effect of one-shot pruning on the YOLOv3-SPP model~\cite{yolov3} for object detection on the COCO dataset~\cite{lin2014microsoft}. We use the state-of-the-art implementation of~\citep{ultralytics}. We one-shot prune this model using global magnitude (the only available baseline) and M-FAC with block size 50K and 1K gradients. (This parameter value would be infeasible for WoodFisher due to storage costs.) 
The results in terms of the standard mAP@$0.5$ metric are provided in Figure~\ref{fig:yolov3}, and show that this model is quite stable under pruning, and that M-FAC provides more accurate pruning for this model as well, relative to magnitude pruning.  

\subsubsection{K-FAC Pruning Comparison}

In Figure~\ref{fig:k-fac-comparison} we compare pruning performance against a pruner which uses K-FAC to estimate second-order statistics (with and without dampening $\pi$) on a fully-connected network and the MNIST dataset. Notice the improved accuracy with M-FAC (and WoodFisher) methods compared to K-FAC, even in the setting with a small network on a simple task.

\subsubsection{Recomputation Effects}

\begin{figure}[h]
    \centering
    \begin{subfigure}[t]{0.45\textwidth}
        \centering
        \includegraphics[width=\linewidth]{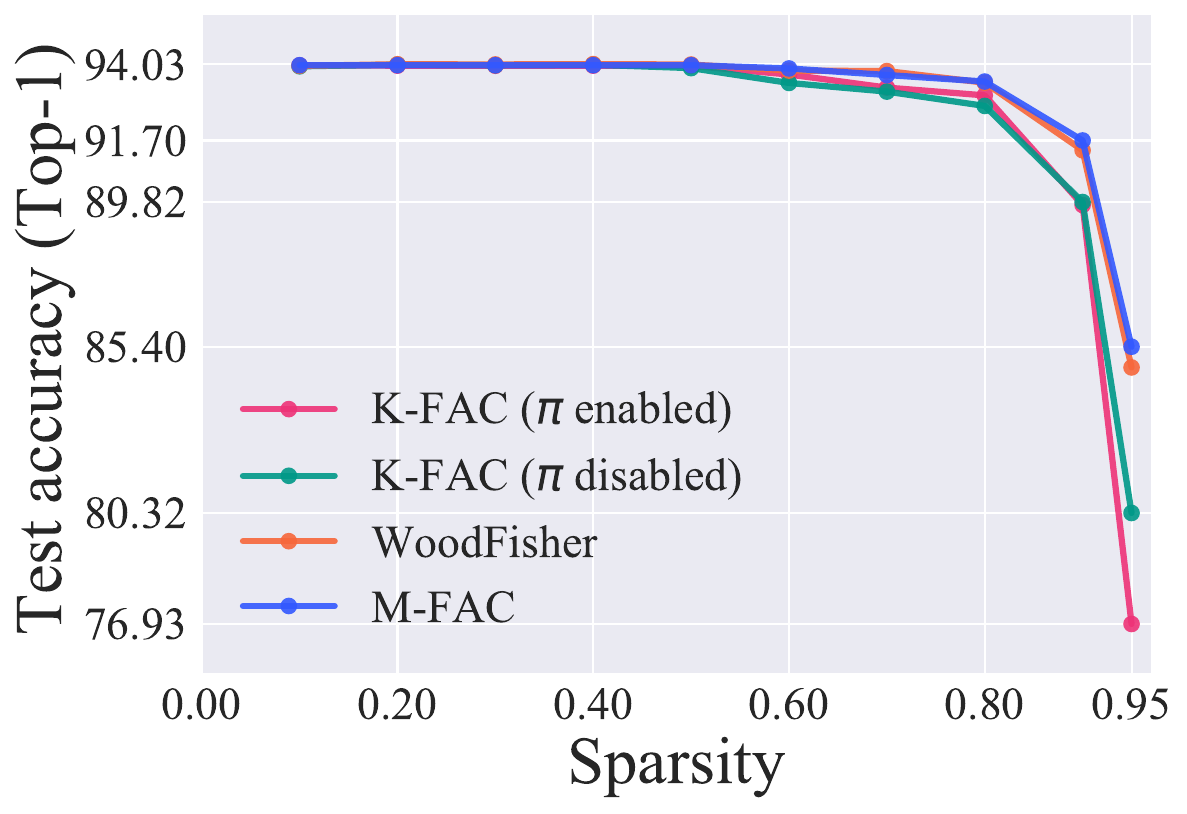}
        \caption{Comparison between K-FAC, WoodFisher and M-FAC in terms of the Top-1 test accuracy for a one-shot pruned Fully-Connected network.}
        \label{fig:k-fac-comparison}
    \end{subfigure}
    \hfill
        \begin{subfigure}[t]{0.45\textwidth}
        \centering
        \includegraphics[width=\linewidth]{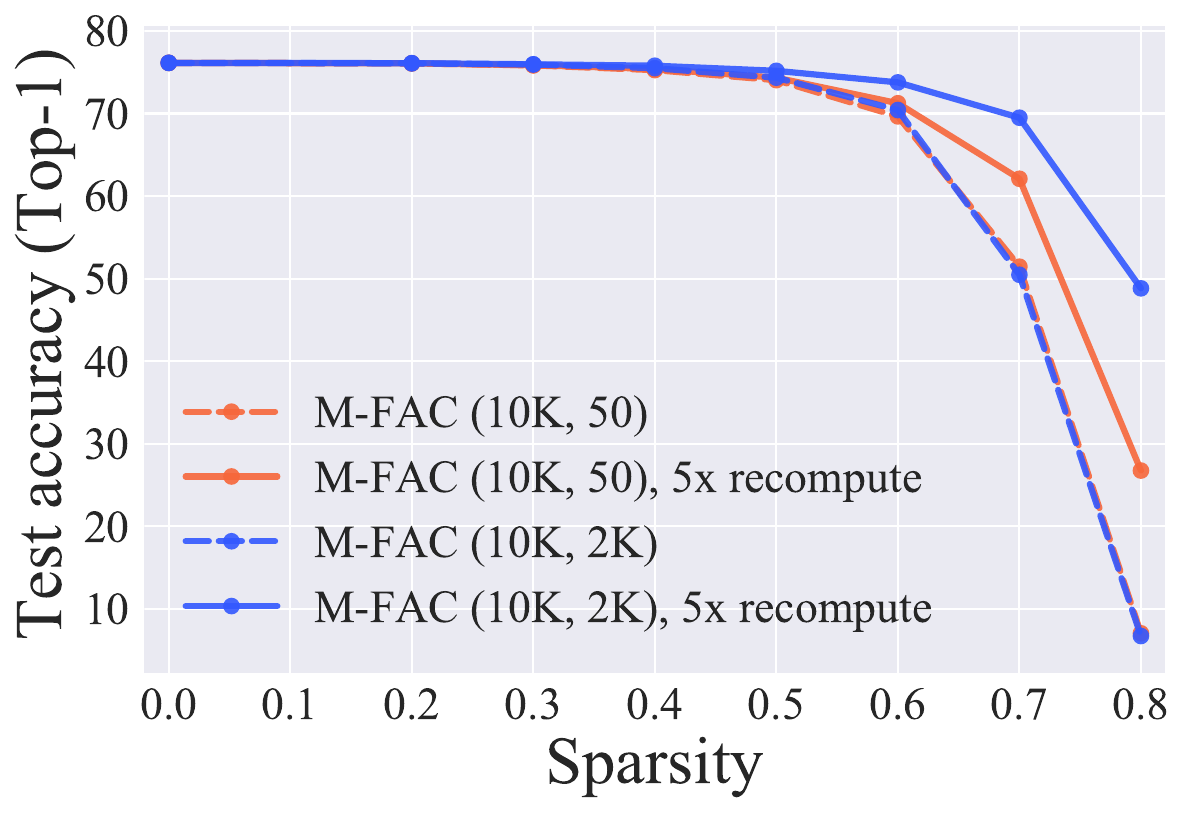}
        \caption{The effect of a recomputation on the accuracy of pruned ResNet50/ImageNet model. Observe the significantly better accuracy provided by recomputation, especially at highwe number of gradients.}
        \label{fig:recomp5}
    \end{subfigure}
    \caption{\textbf{Left}: One-shot pruning comparison for the fully-connected network on the MNIST dataset. \textbf{Right}: The effect of recomputation along the direction of pruning for ResNet50 / ImageNet.}
\end{figure}

As noted before, our one-shot experiments stretch the OBD / OBS theory, as this approach implicitly assumes that the Hessian stays constant across the direction of pruning, which is unlikely to hold for very large pruning displacements. We can however examine the impact of this effect, by \emph{recomputing} the Hessian along the direction of pruning. Figure~\ref{fig:recomp5} shows the effect of recomputation for the ResNet50 / ImageNet model, for 5 recomputation steps, uniformly across the pruning direction. Notice the significant increase in accuracy for the resulting sparse models.

\subsection{Normalization Effects}
\label{sec:oneshot-discussion}

We now discuss our finding that, in the case of some models, e.g. MobileNetV1, smaller blocks appear to yield better accuracy. 
This can be explained by examining the recursive form of the elements of the diagonal inverse. Specifically, without blocking, we get that the $i$th diagonal element has the form
\begin{equation}
    [\widehat{\fisher}^{-1}]_{ii} \simeq \frac{1}{\lambda} \left[ 1 - \frac{\nabla \ell_{1_i}^2}{\| \nabla \ell_1 \|^2 } - \frac{\left(\nabla \ell_{2_i} - \nabla \ell_{1_i} \frac{\nabla \ell_{1}^\top \nabla \ell_2}{\| \nabla \ell_1 \|^2}\right)^2 }{\| \nabla \ell_2 \|^2 - \frac{(\nabla \ell_1^\top \nabla \ell_2)^2}{\| \nabla \ell_1 \|^2} } - \ldots     \right], 
\end{equation}
where $\nabla \ell_{1_i}$ and $\nabla \ell_{2_i}$ represent the $i$th element of a gradient. 
Specifically, notice that, in the case of ``full'' block size (``all'' or $d$), the $i$th gradient entry is divided by the full gradient norm, which may cause it to become negligible if the norm is large. In turn, this leads to essentially the magnitude pruning ranking and update. By contrast, in the case of smaller blocks, the entry is only divided by the norm of the gradient over the block, which mitigates this normalization issue. Similarly, using more gradients also helps, since there are more summands in the above expression, which allows us to deviate from the magnitude pruning baseline.

\subsection{Pruning Experiments Hyperparameters}

\subsubsection{Gradual Pruning Comparisons}

We begin by stating hyperparameters for our gradual pruning runs.
For both MobileNetV1-STR and ResNet50-STR, we employ \emph{identical} hyperparameter values to WoodFisher~\cite{2020-singh}, so that our results are fully comparable. 
The only parameters we change are the ones which pertain to our algorithm, in particular \emph{block size} and \emph{number of gradients} $m$.

For the MobileNetV1-STR gradual experiments, pruning starts from the fully-trained model used by STR / WoodFisher, whose Top-1 validation accuracy is 72\%. 
Gradual pruning is then performed over a total of 24 epochs, pruning additional weights every $4$ epochs, followed by fine-tuning until the next pruning step. 
Pruning targets are set following the standard polynomial schedule of~\cite{2017-zhu, gale2019state}. 
Unless otherwise stated, SGD is used for fine-tuning. 
The learning rate during pruning is $0.005$, momentum is set to $0.9$, and weight decay is $0.0001$. 
Following the last pruning step, fine-tuning is applied until epoch $100$. The initial learning rate value of $0.005$ is decayed multiplicatively by a factor of $0.92$, every epoch, starting from the epoch $30$. 

For ResNet50-STR gradual experiments, pruning starts from the fully-trained model used by STR/WoodFisher, whose Top-1 validation accuracy is 77.01\%. 
Gradual pruning is then performed over a total of $40$ epochs, pruning additional weights every $5$ epochs, followed by fine-tuning until the next pruning step. 
Pruning targets are set following the standard polynomial schedule of~\cite{2017-zhu, gale2019state}. 
Unless otherwise stated, SGD is used for fine-tuning. 
The learning rate during pruning is $0.005$, momentum is set to $0.9$, and weight decay is $0.0001$. 
Following the last pruning step, fine-tuning is applied until epoch $100$. The initial learning rate value of $0.005$ is decayed multiplicatively by a factor of $0.6$, every $6$ epochs, starting from epoch $40$ until epoch $90$.

\subsubsection{Practical Pruning}

In our practical pruning experiments we use a different setup, which is optimized for the particular capabilities of M-FAC. We note that the high speed of M-FAC also makes more careful parameter tuning significantly easier.

In general, we always use M-FAC with block size 128 and 64 gradients (from a batch of 32 samples). Further, each pruning step is performed with 16 recomputations, which always prune the same fraction of remaining weights. This fraction can be calculated as $(i/t)^{1/k}$ where $i$ is the initial sparsity, $t$ is the target sparsity and $k$ is the number of recomputations. We immediately prune down all models to 50\% sparsity before epoch 0 (we found that this sparsity is very easy to recover from, so intermediate steps are unnecessary). Further, we always use SGD without weight decay for a total of 100 epochs and we reset momentum after each pruning step.

For MobileNetV1 we perform a total of 15 additional pruning steps (not counting the initial pruning) which are performed with 3 epochs finetuning in between. All pruning steps prune the same fraction of remaining weights (calculated by the formula described in the previous paragraph). The 3 epochs finetuning in between pruning steps all use the following learning rates $0.005, 0.005, 0.0005$, there is no additional decay (we noticed that dropping the learning rate in the last epoch usually results in a more accurate model as a starting point for the next pruning step). Finally, after the last pruning step at epoch 45, we finetuning for a total of 55 epochs with learning rate $0.005$ with drops by a factor of 10 at epochs 75 and 90.

For Resnet50 we perform a total of 20 additional pruning steps which 4 epochs finetuning at learning rates $0.005, 0.005, 0.005, 0.0005$ in between (due to the higher target sparsities, individual steps are bigger and the model seemed to need more time to recover well in between steps). We first calculate a 15-step equal-fraction schedule to the final target sparsity, execute the first 10 steps and then replace the last 5 with a 10-step equal-fraction schedule from the current to the target sparsity (due to the very high target sparsities the last steps needed to be smaller). Eventually, we finetune for 20 epochs with learning rate $.005$ dropped by a factor of 10 at epochs 90 and 95.

\subsection{Optimization Experiment Hyperparameters}

\paragraph{ResNet20/32}

We now discuss the hyperparameters used for the ResNet20/32 comparison between M-FAC and various first and second order optimizers. Both models are trained with batch-size 128 for 164 epochs ($\approx 64000$ steps) while dropping the learning rate by a factor of $0.1$ after $50\%$ and $75\%$ of training. This is exactly the training setup used by \cite{He_2016}. For M-FAC, we use $m = 512$ gradients, dampening $\lambda = 10^{-5}$ and initial learning rate $10^{-3}$, which were determined to be reasonable default values during development. For the other methods we use tuned initial learning rates from literature; in particular the value for SGD is from \cite{He_2016} while the values for Adam, AdamW and AdaHessian are from \cite{yao2020adahessian}. Since we found that the exact weight decay value can have a significant impact on the final test accuracies, we performed grid-searches over the commonly used values $\{0, 10^{-5}, 10^{-4}, 5 \cdot 10^{-4}, 10^{-3}, 3 \cdot 10^{-3}, 10^{-2}\}$ for all methods that use weight decay. Table \ref{tab:optim-parameters} summarizes the final hyper-parameter settings.

\begin{table}[h]
    \centering
    \begin{tabular}{l|c|c|c}
         Method & learning rate & momentum & weight decay  \\
         \toprule
         SGD & 0.1 & 0.9 & 0.0001 \\
         Adam & 0.001 & (0.9, 0.999) & -- \\
         AdamW & 0.01 & (0.9, 0.999) & 0.01 \\
         AdaHessian & 0.015 & (0.9, 0.999) & 0.003 (RN20), 0.0005 (RN32) \\
         M-FAC ($m = 512$) & 0.001 & -- & 0.003 \\
         \bottomrule
    \end{tabular}
    \caption{Hyperparameter settings used for the ResNet20 (RN20) and ResNet32 (RN32) experiments.}
    \label{tab:optim-parameters}
\end{table}

All experiments were repeated 5 times with different random seeds and we report the median of the best test accuracy; the standard deviations were generally quite low at around $0.2$ percent accuracy.

\paragraph{Hyperparameters for GGT and K-FAC.} Unfortunately, GGT and K-FAC did not produce reasonable results  in the setup discussed so far (i.e. achieved poor accuracy or diverged). Thus, for fairness, we decided to adopt the recommendations of the method authors, even if those were no longer exactly comparable with the other experiments (e.g. different batch-size, learning rate schedule, etc.). Further, we also performed considerably more extensive hyper-parameter searches for these methods. 

For K-FAC, we use the authors' carefully-tuned parameters for ResNet20, which they published in their official repository\footnote{\url{https://github.com/tensorflow/kfac}}. For ResNet32, we adopt the same parameters, but use an initial learning rate of $0.075$ and an initial dampening of $\approx 0.1887$, which we identified to work best for this model via grid search (see Table\ref{tab:kfac-grid} for the grid). The biggest differences of this setup compared to M-FAC are the batch-size of 1000 and the smoothly exponential decaying learning rate.

\begin{table}[h]
    \centering
    \begin{tabular}{l|l}
        \toprule
        Init LR & $0.5, 0.25, 0.1, 0.05, 0.025, 0.01, 0.005, 0.0025, 0.001$ \\
        Final LR & $10^{-4}, 10^{-5}$ \\
        Init Dampening & $0.3887, 0.2887, 0.1887$ \\
        Inversion freq. & $1, 10$ \\
        \bottomrule
    \end{tabular}
    \caption{ResNet32 K-FAC search grid; parameters not listed here were kept the same as the ResNet20 settings.}
    \label{tab:kfac-grid}
\end{table}

For GGT, we use a batch-size of 128 and a gradient window size of $r = 100$ (which is $> 5 \times$ lower than for M-FAC, however we surprisingly found that GGT's performance, unlike M-FAC's, did not improve with larger $r$ and thus kept this as the best value). Further, we use an initial learning rate of $0.05$ and a cosine decaying schedule with $T = 40$, which we found via grid search. All GGT experiments were performed with the author's implementation contributed to TensorFlow. Our GGT search grid is shown in Table \ref{tab:ggt-grid}.

\begin{table}[h]
    \centering
    \begin{tabular}{l|l}
        \toprule
        Init LR & $0.5, 0.1, 0.05, 0.01, 0.005, 0.001, 0.0005, 0.0001$ \\
        Window size $r$ & $50, 100, 200, 500$ \\
        \bottomrule
    \end{tabular}
    \caption{GGT search grid; parameters not listed here were kept at the author's recommended settings.}
    \label{tab:ggt-grid}
\end{table}

\paragraph{Wide ResNet / MobileNetV1.}

Next, we list the exact hyperparameters used for optimizing Wide ResNet (WRN) and MobileNetV1 models. The corresponding experiments were designed to explore how well M-FAC performs relative to other methods without any parameter tuning, i.e. just using reasonable default values. In the case of Wide ResNet, all models are trained for 200 epochs with batch size 256, where the learning rate is dropped by a factor 10 after $50\%$ and $75\%$ of training. SGD uses an initial learning rate of $0.1$ and momentum of $0.9$ (as suggested by \cite{zagoruyko2016wide}). Adam runs with default parameters, i.e. a learning rate of $0.001$, $\beta_1 = 0.9$ and $\beta_2 = 0.999$. M-FAC uses $m = 1024$ and the standard learning rate of $0.001$ (but no momentum). No method uses any weight decay in these experiments. For MobileNetV1, we use exactly the same optimizer settings and learning rate schedule, but we train only for 100 epochs.

\paragraph{Sparse Finetuning.}

For the sparse fine-tuning experiments, pruning occurs identically for all methods. However, we fine-tune the models using either SGD or M-FAC with $m = 512$, for $30$ epochs. Both fine-tuning algorithms are run with exactly the same hyperparameter values: initial learning rate $0.005$, reduced every $10$ epochs by a factor of $10$ and batch-size 256.

\subsection{Wall-Clock Time Comparison.}

\begin{figure}[h]
    \centering
    \begin{subfigure}[t]{0.45\textwidth}
        \centering
        \includegraphics[width=\linewidth]{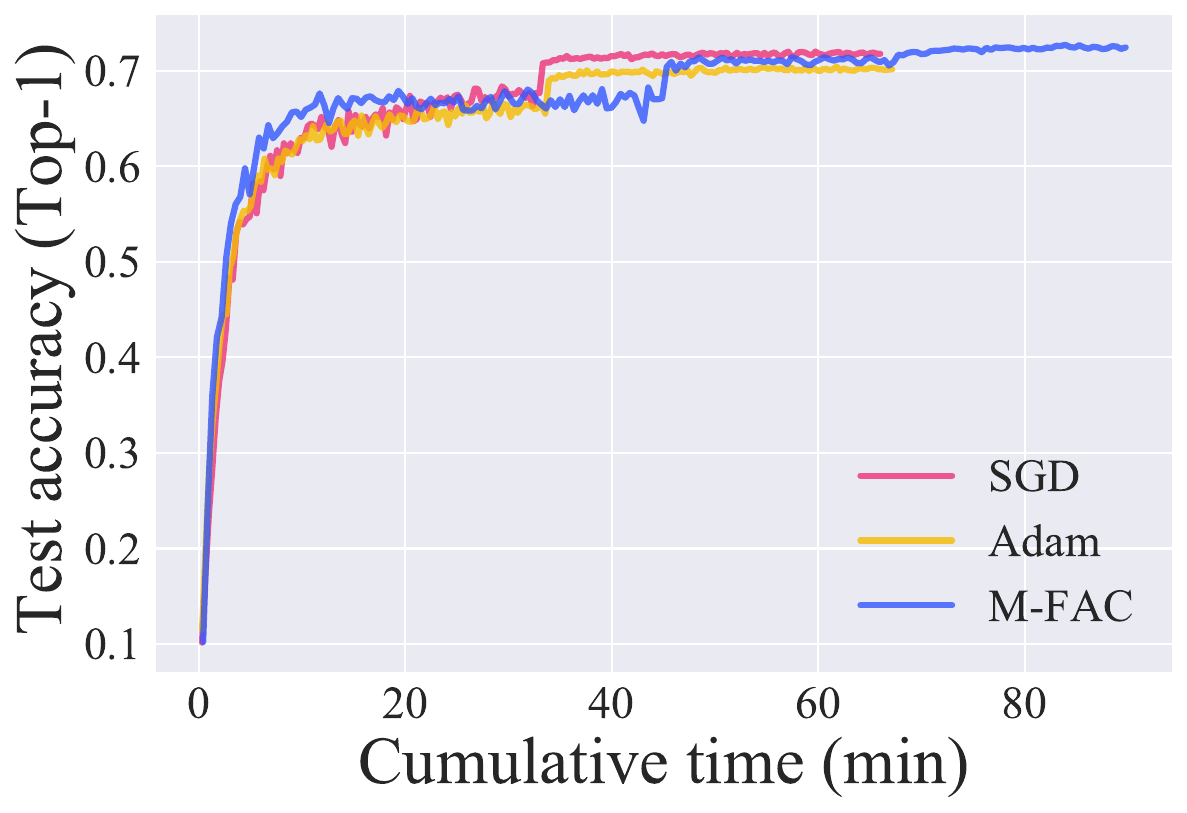}
        \caption{Accuracy versus time for WRN 40-2 / CIFAR-100.}
        \label{fig:wrn40-time}
    \end{subfigure}
    \hfill
    \begin{subfigure}[t]{0.45\textwidth}
        \centering
        \includegraphics[width=\linewidth]{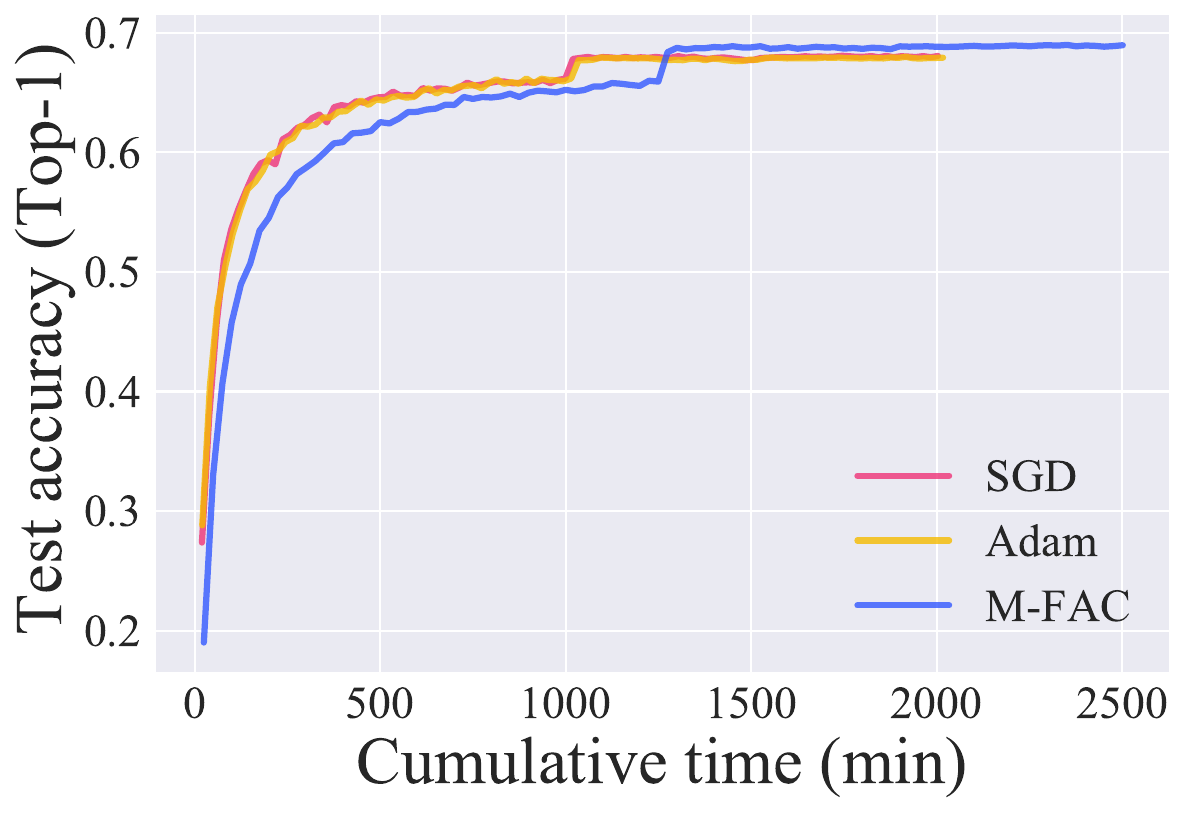}
        \caption{Accuracy versus time for MobileNetV1 / ImageNet.}
        \label{fig:mbv1-time}
    \end{subfigure}
    \caption{Optimization results in accuracy-versus-time format for WRN40-2 and MobileNetV1.}
    \label{fig:mobilenet}
\end{figure}

\begin{figure}[h]
    \centering
    \begin{subfigure}[t]{0.45\textwidth}
        \centering
        \includegraphics[width=\linewidth]{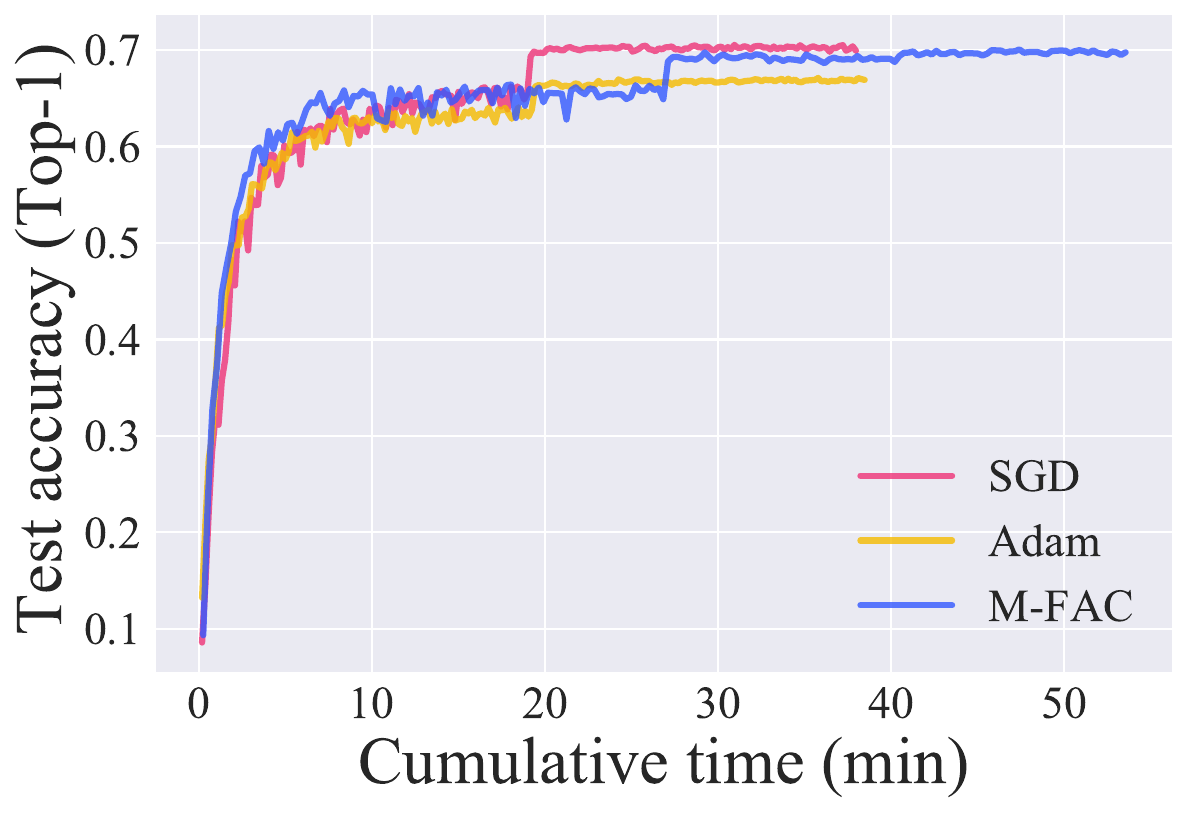}
        \caption{Accuracy versus time for WRN 22-2 / CIFAR-100.}
        \label{fig:wrn22-2-time}
    \end{subfigure}
    \hfill
    \begin{subfigure}[t]{0.45\textwidth}
        \centering
        \includegraphics[width=\linewidth]{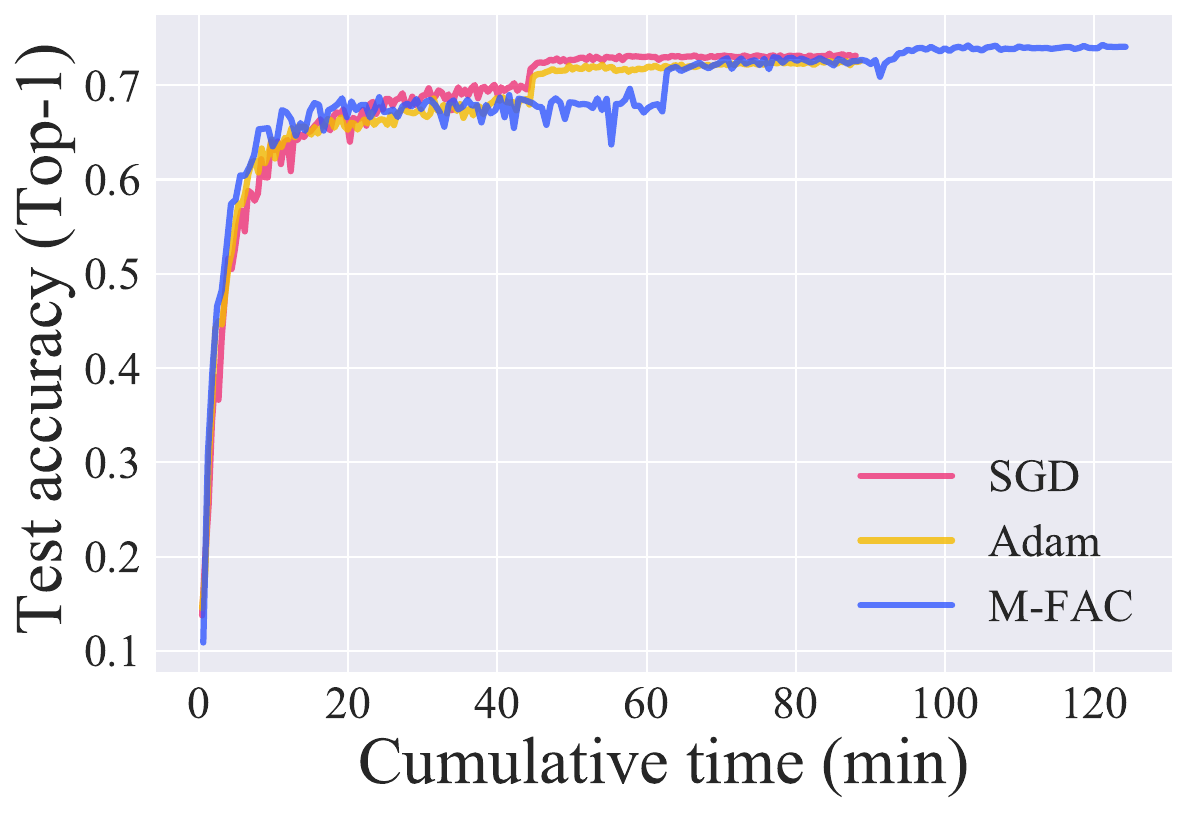}
        \caption{Accuracy versus time for WRN 22-4 / CIFAR-100.}
        \label{fig:wrn224-time}
    \end{subfigure}
    \caption{Optimization results in accuracy-versus-time format for Wide Residual Networks (WRN).}
    \label{fig:wrn}
\end{figure}

In Figures \ref{fig:mobilenet} and \ref{fig:wrn} we show the test accuracies during the training of Wide ResNet (WRN) on the CIFAR-100 dataset and MobileNetV1 on the ImageNet dataset with SGD, Adam and M-FAC with respect to the wall clock time, when executed on a single NVIDIA RTX 3090 GPU. 
(In the case of the largest network, WRN 22-4, we use a second GPU's memory to store gradients, but still use only a single GPU for computation.)
One can see that, in most plots, M-FAC reaches better accuracies than SGD already after only slightly higher training time. Further, we can see that, for the WRN model, the early training accuracy increases fastest for M-FAC even with respect to the wall-clock time, a phenomenon often observed with methods which try to leverage approximate second-order information.


\subsection{Cosine Similarity of Descent Directions}

\begin{figure}[!h]
     \centering
    \includegraphics[width=0.6\linewidth]{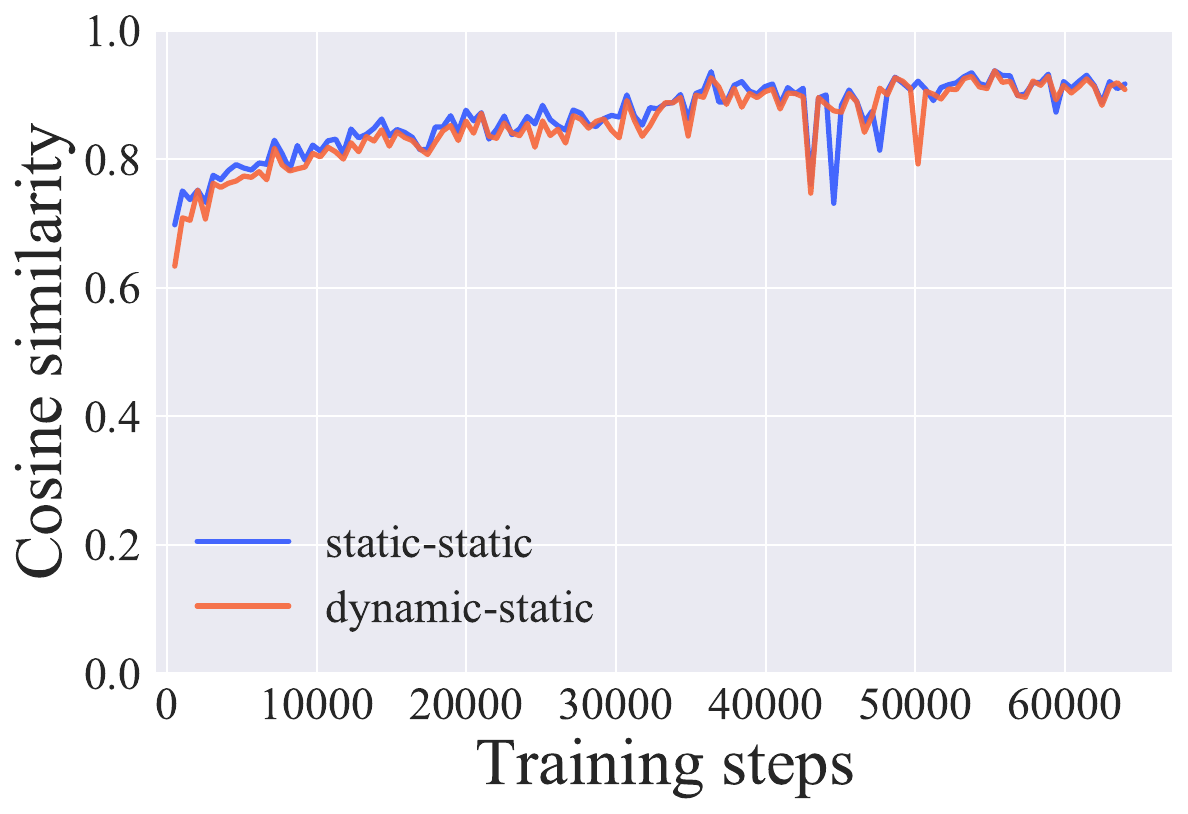}
    \caption{Cosine similarities over the course of training a ResNet20 / CIFAR-10 model. The directions are remarkably close for the dynamic algorithm, suggesting that the sliding window approximation is valid.}
    \label{fig:cosine_sim}
\end{figure}

Finally, in Figure~\ref{fig:cosine_sim}, we examine the quality of the sliding window approximation to the Fisher matrix. We run optimization with M-FAC on ResNet20 / CIFAR-10 (using $m = 512$ and otherwise the same hyperparameters as discussed in the previous section), and every 512 steps we sample $2 \times 512$ gradients to produce 2 static estimates of the Fisher matrix, and then we compare the cosine similarity of the descent direction given by the dynamic algorithm at this step,
$\widehat{F}^{-1}_{\idlow{dynamic}} \cdot \nabla \ell$ with  $\widehat{F}^{-1}_{\idlow{static1}} \cdot \nabla \ell$ (denoted as dynamic--static) as well as the cosine similarity between $\widehat{F}^{-1}_{\idlow{static1}} \cdot \nabla \ell$  and $\widehat{F}^{-1}_{\idlow{static2}} \cdot \nabla \ell$ (denoted as static--static).

The results show that: 1) the cosine similarities between the sliding window approximation and the ``fresh'' approximation are extremely close; 2) they tend to improve significantly as we advance in the optimization process. 
Overall, this validates the sliding window approximation made by our method. 

\section{Transformer Natural Language Modelling}

In this section, we provide more detailed results of our natural language modelling experiments for the tiny (t) and mini (m) variants of the BERT Transformer model. As already stated in the main paper, M-FAC always uses $m = 1024$ gradients (except if a task has significantly less total training batches), dampening $\lambda = 10^{-6}$, learning rate $10^{-4}$ and no weight decay or momentum. We preserve all other configurations (e.g. batchsize, number of epochs, learning rate schedule) from the baseline. The SQuADv2 HuggingFace Adam baseline\footnote{\url{https://github.com/huggingface/transformers/tree/master/examples/pytorch/question-answering}, accessed: 2021-10-26} uses initial learning rate $3 \cdot 10^{-5}$ and trains for 2 epochs, while for the GLUE tasks, it\footnote{\url{https://github.com/huggingface/transformers/tree/master/examples/pytorch/text-classification}, accessed: 2021-10-26} uses initial learning rate $2 \cdot 10^{-5}$ and trains for 3 epochs (5 epochs for MRPC). Table \ref{tab:hf3} and Table \ref{tab:hf4} show our detailed (including all task metrics and standard deviations) question answering SQuADv2 and text-classification GLUE results, respectively. At last, in Table \ref{tab:google}, we compare BERT-tiny models optimized with M-FAC against tuned AdamW results (both trained for 4 epochs) by the BERT authors\footnote{\url{https://github.com/google-research/bert}, accessed: 2021-10-26}, on the GLUE test sets. We find that M-FAC (sometimes with modest tuning) can also outperform this competitive baseline on most tasks, even without using any weight decay or momentum.

\begin{table}[!h]
    \centering
    \begin{tabular}{cl|cc}
        \toprule
         & & SQuADv2 (EM) & SQuADv2 (F1) \\
        \midrule
        t & Adam &  $48.41 \pm 0.57$ & $49.99 \pm 0.54$  \\
        t & M-FAC & \textbf{49.80 $\pm$ 0.43} & \textbf{52.18 $\pm$ 0.20} \\
        \midrule
        m & Adam &  $54.80 \pm 0.47$ & $58.13 \pm 0.31$\\
        m & M-FAC & \textbf{58.02 $\pm$ 0.39} & \textbf{61.35 $\pm$ 0.24} \\
        \bottomrule
    \end{tabular}
    \caption{Comparing the M-FAC optimizer against HuggingFace's Adam baselines on the SQuADv2 dataset.}
    \label{tab:hf3}
\end{table}

\begin{table}[!h]
    \centering
    \begin{tabular}{l|ccccc}
         & SST-2 (Ac) & MRPC (F1) & MRPC (Ac) & STS-B (Pe) & STS-B (Sp) \\
         \toprule
         t / Adam &  $80.11 \pm 0.65$ & $81.68 \pm 0.33$ & $69.90 \pm 0.32$ & $64.39 \pm 5.02$ & $66.52 \pm 5.67$  \\
         t / M-FAC &  \textbf{81.86 $\pm$ 0.76} & \textbf{82.77 $\pm$ 0.22} & \textbf{72.94 $\pm$ 0.37} & \textbf{80.15 $\pm$ 0.52} & \textbf{80.62 $\pm$ 0.43} \\
         \midrule
         m / Adam & \textbf{85.46 $\pm$ 0.58} & $84.57 \pm 0.36$ & $76.57 \pm 0.80$ & $82.09 \pm 0.54$ & $82.64 \pm 0.71$  \\
         m / M-FAC &  $84.20 \pm 0.58$ & \textbf{85.06 $\pm$ 1.63} & \textbf{78.87 $\pm$ 2.33} & \textbf{84.66 $\pm$ 0.30} & \textbf{84.65 $\pm$ 0.30} \\
         \bottomrule
    \end{tabular}
    \newline
    \vspace*{0.5 cm}
    \newline
    \begin{tabular}{l|ccccc}
         & QQP (F1) & QQP (Ac) & MNLI-m (Ac) & MNLI-mm (Ac) & QNLI (Ac) \\
         \toprule
         t / Adam &  $77.58 \pm 0.08$ & $81.09 \pm 0.15$ & $65.36 \pm 0.13$ & $66.78 \pm 0.15$ & $77.85 \pm 0.15$ \\
         t / M-FAC & \textbf{79.71 $\pm$ 0.13} & \textbf{84.29 $\pm$ 0.08} & \textbf{68.28 $\pm$ 3.29} & \textbf{68.98 $\pm$ 3.05} & \textbf{81.17 $\pm$ 0.43} \\
         \midrule
         m / Adam &  $82.43 \pm 0.10$ & $86.45 \pm 0.12$ & $73.30 \pm 0.20$ & $74.85 \pm 0.09$ & \textbf{83.85 $\pm$ 0.10} \\
         m / M-FAC &  \textbf{82.67 $\pm$ 0.23} & \textbf{86.75 $\pm$ 0.20} & \textbf{74.59 $\pm$ 0.41} & \textbf{75.95 $\pm$ 0.14} & $83.70 \pm 0.13$ \\
         \bottomrule
    \end{tabular}
    \caption{Comparing the M-FAC optimizer against HuggingFace's Adam baselines on the GLUE benchmark suite (we show scores on the validation set).}
    \label{tab:hf4}
\end{table}

\begin{table}[!h]
    \centering
    \begin{tabular}{l|ccccc}
        \toprule
        & SST-2 (Ac) & MRPC (F1) & MRPC (Ac) & STS-B (Pe) & STS-B (Sp) \\
        \midrule
        AdamW & $83.2$ & $81.1$ & $71.1$ & $74.3$ & \textbf{73.6} \\
        M-FAC & \textbf{83.4$^*$} & \textbf{81.9$^*$} & \textbf{72.7$^*$} & \textbf{75.3$^*$} & $73.2^*$  \\
        \bottomrule
    \end{tabular}
    \newline
    \vspace*{0.5 cm}
    \newline
    \begin{tabular}{l|ccccc}
        \toprule
        & QQP (F1) & QQP (Ac) & MNLI-m (Ac) & MNLI-mm (Ac) & QNLI (Ac) \\
        \midrule
        AdamW & $62.2$ & $83.4$ & $70.2$ & $70.3$ & $81.5$\\
        M-FAC & \textbf{62.8} & \textbf{83.9} & \textbf{71.0} & \textbf{70.5} & \textbf{81.7} \\
        \bottomrule
    \end{tabular}
    \caption{Comparing M-FAC optimizer (without weight decay) against BERT authors' \textit{tuned} BERT-tiny AdamW baseline on a subset of GLUE benchmark \textit{test} sets. $^*$ Modest tuning of learning rate and dampening because of the very low number of samples / gradients in the training data.}
    \label{tab:google}
\end{table}

\end{document}